\documentclass[letterpaper, 10pt, conference]{ieeeconf}

\IEEEoverridecommandlockouts%
\usepackage{amsmath,amssymb}
\usepackage{mathtools}%
\usepackage{bm}
\usepackage{siunitx}

\usepackage{graphicx}
\graphicspath{{figures/}}
\usepackage[caption=false,font=footnotesize]{subfig}

\usepackage{capt-of}
\usepackage{booktabs}
\usepackage{multirow}
\usepackage{pifont}%
\usepackage{xcolor}
\usepackage{tikz}
\usetikzlibrary{arrows.meta,positioning,calc,fit,shapes.geometric,shapes.callouts,decorations.pathreplacing,backgrounds}

\usepackage{xspace}
\usepackage{acro}%
\acsetup{make-links = true}
\DeclareAcronym{vla}{short = VLA, long = vision-language-action}
\DeclareAcronym{vlm}{short = VLM, long = vision-language model}
\DeclareAcronym{llm}{short = LLM, long = large language model}
\DeclareAcronym{ml} {short = ML,  long = machine learning}
\DeclareAcronym{ai} {short = AI,  long = artificial intelligence}
\DeclareAcronym{ood}{short = OOD, long = out-of-distribution}
\DeclareAcronym{sr} {short = SR,  long = success rate}

\DeclareAcronym{cctv}{short = CCTV, long = closed-circuit television}
\DeclareAcronym{bev} {short = BEV,  long = bird's-eye view}
\DeclareAcronym{cnn} {short = CNN,  long = convolutional neural network}
\DeclareAcronym{nlp} {short = NLP,  long = natural language processing}
\DeclareAcronym{rl}  {short = RL,   long = reinforcement learning}
\DeclareAcronym{iot} {short = IoT,  long = Internet of Things}
\DeclareAcronym{fmm} {short = FMM,  long = fast marching method}

\DeclareAcronym{mlp} {short = MLP,  long = multilayer perceptron}
\DeclareAcronym{vit} {short = ViT,  long = vision transformer}

\DeclareAcronym{edf} {short = EDF,  long = Euclidean distance field}

\usepackage[hidelinks]{hyperref}
\hypersetup{pdfauthor={Lukas Vierling, Benjamin Ramtoula, Luke Robinson, Ronald Clark, Daniele De Martini}, pdftitle={InfraVLA: Extending Vision-Language-Action Navigation with Infrastructure Cameras}, pdfsubject={}, pdfkeywords={}}
\usepackage[capitalise]{cleveref}%
\Crefname{figure}{Fig.}{Figs.}
\crefformat{equation}{(#2#1#3)}
\crefrangeformat{equation}{(#3#1#4) to (#5#2#6)}
\crefmultiformat{equation}{(#2#1#3)}{ and (#2#1#3)}{, (#2#1#3)}{, and (#2#1#3)}

\definecolor{markyes}{HTML}{1E7B45}
\definecolor{markno}{HTML}{C0504D}

\newcommand{\E}{\mathbb{E}}

\DeclarePairedDelimiter{\norm}{\lVert}{\rVert}

\newcommand{\astar}{A*\xspace}
\newcommand{\method}{InfraVLA\xspace}%

\title{\LARGE \bf
InfraVLA: Extending Vision-Language-Action Navigation with Infrastructure Cameras
}

\author{Lukas Vierling$^{1}$, Benjamin Ramtoula$^{2*}$, Luke Robinson$^{2}$, Ronald Clark$^{1}$ and Daniele De Martini$^{2}$\\[4pt]
{\small $^{1}$Dept.\ of Computer Science, $^{2}$Oxford Robotics Institute, University of Oxford \quad $^{*}$\texttt{benjamin@robots.ox.ac.uk}}}

\makeatletter
\def\bstctlcite{\@ifnextchar[{\@bstctlcite}{\@bstctlcite[@auxout]}}
\def\@bstctlcite[#1]#2{\@bsphack
  \@for\@citeb:=#2\do{%
    \edef\@citeb{\expandafter\@firstofone\@citeb}%
    \if@filesw\immediate\write\csname #1\endcsname{\string\citation{\@citeb}}\fi}%
  \@esphack}
\makeatother
\begin{document}
\bstctlcite{IEEEexample:BSTcontrol}

\IEEEaftertitletext{%
  \vspace{1.0mm}%
  \noindent\begin{minipage}{0.9\textwidth}%
    \centering
    \definecolor{hrOrange}{RGB}{230,159,0}%
\definecolor{hrGreen}{RGB}{0,158,115}%
\definecolor{hrBlue}{RGB}{0,114,178}%
\definecolor{hrVerm}{RGB}{213,94,0}%
\definecolor{hrPurple}{RGB}{204,121,167}%
\definecolor{hrSky}{RGB}{86,180,233}%
\definecolor{hrTan}{RGB}{150,100,60}%
\definecolor{hrInk}{RGB}{26,26,26}%

\providecommand{\heroW}{8.55cm}%
\providecommand{\heroHalf}{4.215cm}%
\providecommand{\heroWW}{17.30cm}%

\tikzset{
  hrtag/.style={font=\sffamily\scriptsize, text=hrInk, inner sep=0.55mm,
                rounded corners=0.4mm, fill=white, fill opacity=0.84,
                text opacity=1, align=center},
  hrtagsm/.style={hrtag, font=\sffamily\fontsize{6}{7}\selectfont},
  hrhead/.style={font=\sffamily\footnotesize, text=black!80, align=left,
                 inner sep=0pt},
  hrheadsm/.style={font=\sffamily\scriptsize, text=black!75, align=left,
                   inner sep=0pt},
  hrchip/.style={font=\sffamily\footnotesize, text=hrInk, align=left,
                 inner sep=0.9mm, rounded corners=0.9mm,
                 draw=hrBlue!70!black, line width=0.5pt, fill=hrBlue!10},
  hrring/.style={circle, draw=#1, line width=0.85pt, inner sep=0pt,
                 minimum size=4.6mm},
  hrring/.default=hrBlue,
  hrhalo/.style={circle, draw=white, line width=1.9pt, inner sep=0pt,
                 minimum size=4.6mm, opacity=0.55},
  hrringsm/.style={circle, draw=#1, line width=0.7pt, inner sep=0pt,
                   minimum size=3.0mm},
  hrringsm/.default=hrBlue,
  hrhalosm/.style={circle, draw=white, line width=1.5pt, inner sep=0pt,
                   minimum size=3.0mm, opacity=0.55},
  hrlead/.style={draw=white, line width=1.6pt, opacity=0.6},
  hrleadink/.style={draw=hrInk, line width=0.5pt},
  hrshelf/.style={fill=black!62, draw=none},
  hrwall/.style={draw=black!55, line width=0.7pt, fill=white},
  hrroute/.style={-{Stealth[length=1.5mm,width=1.2mm]}, line width=0.7pt,
                  rounded corners=1.2mm},
  hrcand/.style={hrroute, draw=black!38, densely dashed},
  hrbox/.style={rounded corners=0.7mm, draw=#1!70!black, fill=#1!14,
                font=\sffamily\scriptsize, text=hrInk, align=center,
                inner sep=0.7mm},
  hrmodel/.style={rounded corners=0.9mm, draw=black!60, fill=black!10,
                  font=\sffamily\scriptsize, text=hrInk, align=center,
                  inner sep=0.8mm},
  hrflow/.style={-{Stealth[length=1.6mm,width=1.3mm]}, draw=black!55,
                 line width=0.6pt},
}

\providecommand{\heroRel}[2]{%
  \begin{scope}[shift={(#1.south west)},
                x={($(#1.south east)-(#1.south west)$)},
                y={($(#1.north west)-(#1.south west)$)}]%
    #2%
  \end{scope}}

\providecommand{\heroMark}[3][hrBlue]{%
  \node[hrhalo] at (#2,#3) {};%
  \node[hrring=#1] (hrlast) at (#2,#3) {};}

\providecommand{\heroMarkS}[3][hrBlue]{%
  \node[hrhalosm] at (#2,#3) {};%
  \node[hrringsm=#1] (hrlasts) at (#2,#3) {};}

\providecommand{\heroFrame}[3]{%
  \setlength{\fboxsep}{0pt}\setlength{\fboxrule}{0.4pt}%
  \textcolor{black!45}{\fbox{\includegraphics[width=#1, trim=#2, clip]{#3}}}}

\providecommand{\hrRobot}[3]{%
  \begin{scope}[shift={(#1,#2)}, scale=#3]
    \fill[black!78, rounded corners=0.3mm] (-0.20,-0.13) rectangle (0.20,0.13);
    \fill[hrSky] (0.20,-0.05) rectangle (0.27,0.05);
  \end{scope}}
\providecommand{\hrCam}[4]{%
  \begin{scope}[shift={(#1,#2)}, rotate=#3, scale=#4]
    \fill[black!80] (-0.06,-0.09) -- (0.20,-0.13) -- (0.20,0.13) -- (-0.06,0.09) -- cycle;
    \fill[black!80] (-0.16,0.00) circle (0.045);
  \end{scope}}
\providecommand{\hrCrate}[3]{%
  \fill[hrTan, draw=black!55, line width=0.3pt]
    ($(#1,#2)+(-0.10,-0.10)$) rectangle ($(#1,#2)+(0.10,0.10)$);}
\providecommand{\hrBarrel}[3]{%
  \fill[hrBlue, draw=black!55, line width=0.3pt] (#1,#2) circle (0.11);}
\providecommand{\hrCone}[3]{%
  \fill[hrVerm, draw=black!55, line width=0.3pt]
    ($(#1,#2)+(-0.10,-0.09)$) -- ($(#1,#2)+(0.10,-0.09)$) -- ($(#1,#2)+(0,0.11)$) -- cycle;}
\providecommand{\hrFork}[3]{%
  \fill[yellow!85!orange, draw=black!55, line width=0.3pt]
    ($(#1,#2)+(-0.16,-0.09)$) rectangle ($(#1,#2)+(0.16,0.09)$);}

\providecommand{\heroFrameC}[4]{%
  \setlength{\fboxsep}{0pt}\setlength{\fboxrule}{0.7pt}%
  \textcolor{#1}{\fbox{\includegraphics[width=#2, trim=#3, clip]{#4}}}}

\providecommand{\heroFade}[2]{%
  \heroRel{#1}{\fill[white, opacity=#2] (0,0) rectangle (1,1);}}

\providecommand{\heroCross}[1]{%
  \node[circle, fill=white, draw=black!45, line width=0.4pt, inner sep=0pt,
        minimum size=2.6mm, font=\sffamily\bfseries\fontsize{6}{6}\selectfont,
        text=black!60] at #1 {$\times$};}

\tikzset{
  hrok/.style={circle, fill=hrGreen, draw=white, line width=0.6pt, text=white,
               font=\bfseries\scriptsize, inner sep=0pt, minimum size=3.7mm},
  hrbad/.style={circle, fill=hrVerm, draw=white, line width=0.6pt, text=white,
                font=\bfseries\scriptsize, inner sep=0pt, minimum size=3.7mm},
  hrstamp/.style={font=\sffamily\scriptsize, text=black!55, inner sep=0.8mm,
                  rounded corners=0.5mm, fill=white, fill opacity=0.85,
                  text opacity=1, align=center},
}
\providecommand{\heroOK}[2]{\node[hrok] at (#1,#2) {$\checkmark$};}
\providecommand{\heroBad}[2]{\node[hrbad] at (#1,#2) {$\times$};}
\providecommand{\heroTagTR}[1]{\node[hrtagsm, anchor=north east] at (0.975,0.955) {#1};}

\providecommand{\heroGlyph}[4]{%
  \foreach \dx/\dy in {-0.35pt/0pt, 0.35pt/0pt, 0pt/-0.35pt, 0pt/0.35pt,
                       -0.25pt/-0.25pt, 0.25pt/0.25pt, -0.25pt/0.25pt, 0.25pt/-0.25pt}{
    \node[text=white, font=\Large, inner sep=0pt, xshift=\dx, yshift=\dy]
      at (#1,#2) {#4};}
  \node[text=#3, font=\Large, inner sep=0pt] at (#1,#2) {#4};}
\providecommand{\heroTick}[2]{\heroGlyph{#1}{#2}{hrGreen!85!black}{\ding{51}}}
\providecommand{\heroWrong}[2]{\heroGlyph{#1}{#2}{hrVerm}{\ding{55}}}

\providecommand{\heroFrameCW}[5]{%
  \setlength{\fboxsep}{0pt}\setlength{\fboxrule}{#2}%
  \textcolor{#1}{\fbox{\includegraphics[width=#3, trim=#4, clip]{#5}}}}

\providecommand{\heroMarkD}[3][hrSky]{%
  \node[circle, draw=black!78, line width=1.7pt, inner sep=0pt,
        minimum size=3.2mm] at (#2,#3) {};%
  \node[circle, draw=#1, line width=0.8pt, inner sep=0pt,
        minimum size=3.2mm] at (#2,#3) {};}

\providecommand{\heroPlan}[1]{%
  \setlength{\fboxsep}{0pt}\setlength{\fboxrule}{0.4pt}%
  \textcolor{black!45}{\fbox{\includegraphics[width=#1,
    trim=30.3bp 60.6bp 5.9bp 26.2bp, clip]{sub_layout.pdf}}}}

\tikzset{
  hrobj/.style={draw=black!60, line width=0.3pt, inner sep=0pt, fill=#1,
                minimum size=1.7mm},
  hrcamdot/.style={circle, draw=white, line width=0.5pt, fill=#1, inner sep=0pt,
                   minimum size=1.7mm},
  hrbot/.style={rectangle, rounded corners=0.25mm, draw=white, line width=0.4pt,
                fill=black!80, inner sep=0pt, minimum width=2.2mm,
                minimum height=1.5mm},
}
\providecommand{\hrBox}[2]{\node[hrobj=hrTan] at (#1,#2) {};}
\providecommand{\hrBar}[2]{\node[hrobj=hrBlue, circle] at (#1,#2) {};}
\providecommand{\hrCn}[2]{\node[hrobj=hrVerm, isosceles triangle,
  isosceles triangle apex angle=70, rotate=90, minimum size=1.5mm] at (#1,#2) {};}
\providecommand{\hrFk}[2]{\node[hrobj=yellow!85!orange, minimum width=2.6mm,
  minimum height=1.5mm] at (#1,#2) {};}
\providecommand{\hrBotAt}[2]{\node[hrbot] at (#1,#2) {};}
\providecommand{\hrCams}{%
  \node[hrcamdot=hrBlue]   at (0.22,0.965) {};
  \node[hrcamdot=hrGreen]  at (0.81,0.965) {};
  \node[hrcamdot=hrPurple] at (0.22,0.035) {};
  \node[hrcamdot=hrVerm]   at (0.81,0.035) {};}
\providecommand{\hrObjs}{%
  \hrFk{0.279}{0.81}\hrBar{0.645}{0.81}\hrBox{0.279}{0.20}\hrCn{0.645}{0.20}}
\providecommand{\hrSpawn}{\hrBotAt{0.10}{0.507}}
\providecommand{\hrRouteB}[1]{%
  \draw[hrroute, draw=#1, line width=0.9pt]
    (0.115,0.507) .. controls (0.24,0.515) and (0.279,0.58) .. (0.279,0.715);}
\providecommand{\hrRouteC}[1]{%
  \draw[hrroute, draw=#1, line width=0.9pt]
    (0.115,0.507) .. controls (0.50,0.515) and (0.645,0.58) .. (0.645,0.715);}

\providecommand{\heroTokens}[2]{%
  \begin{scope}[shift={#1}, scale=#2]
    \node[hrbox=hrOrange, anchor=west]  (t1) at (0,1.05) {on-board};
    \node[hrbox=hrGreen, anchor=west]   (t2) at ($(t1.east)+(0.09,0)$) {CCTV $\times N$};
    \node[hrbox=hrBlue, anchor=west]    (t3) at ($(t2.east)+(0.09,0)$) {instruction};
    \node[hrmodel, anchor=north, minimum height=4.2mm]
      (bb) at ($(t2.south)+(0,-0.34)$) {VLA backbone};
    \node[hrbox=hrPurple, anchor=west]  (wp) at ($(bb.east)+(0.14,0)$) {waypoints};
    \draw[hrflow] (t1.south) -- ($(bb.north west)+(0.05,0)$);
    \draw[hrflow] (t2.south) -- (bb.north);
    \draw[hrflow] (t3.south) -- ($(bb.north east)+(-0.05,0)$);
    \draw[hrflow] (bb.east) -- (wp.west);
  \end{scope}}

\providecommand{\heroData}[2]{%
  \begin{scope}[shift={#1}, scale=#2]
    \fill[black!70] (0,0) circle (0.045);
    \draw[hrroute, draw=hrBlue] (0,0) .. controls (0.45,0.02) .. (0.80,0.34);
    \draw[hrroute, draw=hrVerm] (0,0) .. controls (0.45,-0.02) .. (0.80,-0.34);
    \node[hrtagsm, anchor=west] at (0.84,0.34) {instruction A};
    \node[hrtagsm, anchor=west] at (0.84,-0.34) {instruction B};
    \node[hrheadsm, anchor=north west] at (-0.06,-0.52) {counterfactual pair};
    \begin{scope}[shift={(2.55,0)}]
      \draw[line width=0.7pt, black!30] (-0.15,-0.30) -- (1.05,-0.30);
      \draw[hrroute, draw=black!45, densely dashed]
        (-0.05,-0.30) .. controls (0.35,-0.24) .. (0.55,0.10);
      \draw[hrroute, draw=hrGreen] (0.55,0.10) .. controls (0.80,0.05) .. (0.95,-0.26);
      \node[hrtagsm, anchor=south west] at (0.30,0.12) {drift};
      \node[hrtagsm, anchor=west] at (0.99,-0.20) {return};
      \node[hrheadsm, anchor=north west] at (-0.15,-0.52) {recovery data};
    \end{scope}
  \end{scope}}
\providecommand{\method}{InfraVLA}%
\tikzset{
  hxtile/.style={rectangle, draw=#1!75!black, fill=#1!45, inner sep=0pt,
                 minimum width=2.2mm, minimum height=2.2mm, line width=0.35pt},
  hxseqbox/.style={rounded corners=0.9mm, draw=black!35, fill=black!4, line width=0.5pt},
  hxmodel/.style={rounded corners=1.1mm, draw=hrBlue!75!black, fill=hrBlue!12,
                  line width=0.7pt, align=center, text=hrInk, font=\sffamily\small},
  hxflow/.style={-{Stealth[length=1.8mm,width=1.5mm]}, draw=black!50, line width=0.65pt},
  hxflowthick/.style={hxflow, line width=0.95pt, draw=black!60},
  hxtag/.style={font=\sffamily\fontsize{6}{7}\selectfont, text=hrInk, inner sep=0.5mm,
                rounded corners=0.4mm, fill=white, fill opacity=0.9, text opacity=1},
  hxtagc/.style={hxtag, text=#1, draw=#1, line width=0.3pt},
  hxpath/.style={draw=hrGreen!80!black, line width=0.95pt, rounded corners=0.5mm},
  hxpathhalo/.style={draw=white, line width=2.3pt, opacity=0.95, rounded corners=0.5mm},
  hxhead/.style={font=\sffamily\fontsize{6.5}{7}\selectfont, text=black!60, anchor=south west, inner sep=0pt},
  hxnum/.style={circle, fill=black!55, text=white, font=\sffamily\bfseries\fontsize{5}{5}\selectfont,
                inner sep=0pt, minimum size=2.6mm, anchor=south west},
}
\providecommand{\hxHead}[4]{%
  \node[hxnum, anchor=center] (hxn) at (#1+0.13,#2+0.085) {#3};
  \node[hxhead, anchor=base west] at (#1+0.34,#2) {#4};}
\providecommand{\hxGroupN}[4]{%
  \foreach \i in {1,...,#4}{\node[hxtile=#3] at ($#2+(0,{((#4+1)/2-\i)*2.75mm})$) {};}%
  \foreach \k in {0,1,2}{\fill[#3!65!black]
    ($#2+(0,{-((#4-1)/2*2.75mm+1.1mm+1.0mm+\k*0.85mm)})$) circle (0.26mm);}}
\providecommand{\hxIconSpeech}[2]{%
  \begin{scope}[shift={(#1,#2)}]
    \fill[hrBlue!80!black, rounded corners=0.45mm] (-0.25,-0.11) rectangle (0.25,0.16);
    \fill[hrBlue!80!black] (-0.14,-0.10) -- (0.00,-0.10) -- (-0.15,-0.24) -- cycle;
    \foreach \x in {-0.11,0,0.11}{\fill[white] (\x,0.025) circle (0.028);}
  \end{scope}}
\providecommand{\hxIconCCTV}[2]{%
  \begin{scope}[shift={(#1,#2)}]
    \fill[hrGreen!65!black] (-0.20,0.19) rectangle (0.20,0.235);%
    \fill[hrGreen!65!black] (-0.03,0.05) rectangle (0.03,0.20);%
    \begin{scope}[rotate=-22]
      \fill[hrGreen!65!black, rounded corners=0.3mm] (-0.24,-0.09) rectangle (0.16,0.09);
      \fill[hrGreen!65!black] (0.16,-0.12) -- (0.26,-0.14) -- (0.26,0.14) -- (0.16,0.12) -- cycle;
      \fill[white] (0.215,0) circle (0.035);
      \draw[hrGreen!65!black, line width=0.35pt] (0.28,0.05) -- (0.40,0.125);
      \draw[hrGreen!65!black, line width=0.35pt] (0.28,-0.05) -- (0.40,-0.125);
    \end{scope}
  \end{scope}}
\providecommand{\hxIconRobot}[2]{%
  \begin{scope}[shift={(#1,#2)}]
    \fill[hrOrange!80!black, rounded corners=0.4mm] (-0.27,-0.09) rectangle (0.27,0.10);
    \fill[hrOrange!80!black] (-0.16,-0.15) circle (0.065);
    \fill[hrOrange!80!black] (0.16,-0.15) circle (0.065);
    \fill[hrOrange!80!black, rounded corners=0.2mm] (0.10,0.10) rectangle (0.24,0.19);
    \fill[white] (0.235,0.145) circle (0.025);
    \draw[hrOrange!80!black, line width=0.35pt] (0.28,0.145) -- (0.40,0.225);
    \draw[hrOrange!80!black, line width=0.35pt] (0.28,0.145) -- (0.40,0.065);
  \end{scope}}
\providecommand{\heroTrackFile}{figures/hero/proposals/track_v4_b02}%
\providecommand{\heroCctvTrim}{0 389 0 0}%
\providecommand{\heroObTrim}{58 96 58 106}%
\providecommand{\heroSceneTrim}{0 115 0 288}%
\providecommand{\heroCctvFile}{hero/sim_views/run19_v4_clean/cctv_NE.png}%
\providecommand{\heroCctvBoxX}{0.546}\providecommand{\heroCctvBoxY}{0.397}%
\providecommand{\heroObFile}{hero/sim_views/rollout_frames/v4_onboard_start.png}%
\providecommand{\heroSceneFile}{hero/sim_views/baked/v4_b02_path.png}%
\providecommand{\heroObPathFile}{hero/sim_views/baked/v4_onboard_path.png}%
\providecommand{\heroSceneBoxX}{0.609}\providecommand{\heroSceneBoxY}{0.828}%
\providecommand{\heroSceneOverlay}{\heroMarkS[hrBlue]{\heroSceneBoxX}{\heroSceneBoxY}}%
\def\ppTrackHero{(0.1634,0.2415) (0.1925,0.2211) (0.2280,0.1969) (0.2653,0.1752) (0.3036,0.1618) (0.3456,0.1607) (0.3908,0.1723) (0.4356,0.1929) (0.4780,0.2187) (0.5158,0.2478) (0.5480,0.2806) (0.5743,0.3181) (0.5944,0.3587) (0.6092,0.3996) (0.6193,0.4387) (0.6259,0.4755) (0.6302,0.5096) (0.6327,0.5408) (0.6343,0.5693) (0.6348,0.5915) (0.6345,0.6078) (0.6335,0.6200) (0.6320,0.6327) (0.6314,0.6369)}
\def\ppStartHero{(0.1573,0.2575)}\def\ppStartHeroX{0.1573}\def\ppStartHeroY{0.2575}
\def\ppEndHero{(0.6328,0.6504)}\def\ppEndHeroX{0.6328}\def\ppEndHeroY{0.6504}
\def\ppBoxHero{(0.6090,0.6758)}\def\ppBoxHeroX{0.6090}\def\ppBoxHeroY{0.6758}
\begin{tikzpicture}[x=1cm, y=1cm, font=\sffamily]
  \useasboundingbox (0,0.25) rectangle (17.30,4.60);
  \def\ctop{4.05}
  \hxHead{0.75}{4.22}{1}{Inputs}
  \hxHead{4.35}{4.22}{2}{Tokens}
  \hxHead{5.67}{4.22}{3}{Policy}
  \hxHead{8.28}{4.22}{4}{Robot finds the box}
  \def\xin{0.75}\def\win{2.85}
  \node[hrchip, anchor=north west, minimum width=\win cm, minimum height=5.2mm,
        align=center] (chip) at (\xin,\ctop) {``move toward the box''};
  \node[inner sep=0pt, anchor=north west] (cctv) at (\xin,3.37)
    {\edef\hxtmp{\noexpand\heroFrameC{hrGreen}{\win cm}{\heroCctvTrim}{\heroCctvFile}}\hxtmp};
  \node[inner sep=0pt, anchor=north west] (ob) at (\xin,1.90)
    {\edef\hxtmp{\noexpand\heroFrameC{hrOrange}{\win cm}{\heroObTrim}{\heroObFile}}\hxtmp};
  \heroRel{cctv}{%
    \heroTagTR{ceiling camera}%
    \heroMarkS[hrBlue]{\heroCctvBoxX}{\heroCctvBoxY}%
    \node[hxtagc=hrBlue!85!black, anchor=east] at ({\heroCctvBoxX-0.055},\heroCctvBoxY) {target visible};}
  \heroRel{ob}{%
    \heroTagTR{on-board camera}%
    \node[hxtag, anchor=south west, text=black!60] at (0.025,0.05) {target not visible};}
  \hxIconSpeech{0.30}{3.79}
  \node[inner sep=0pt, anchor=center] at (0.30,2.715) {\includegraphics[width=5.8mm]{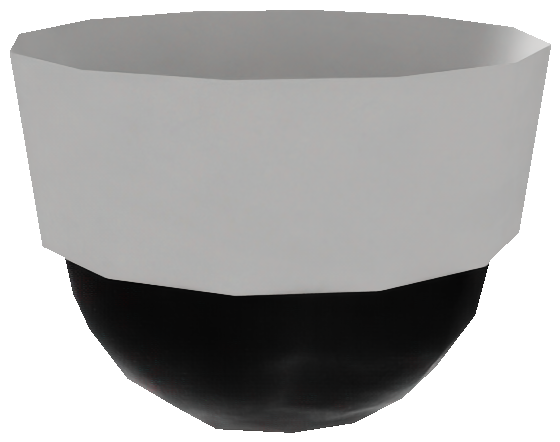}};
  \node[inner sep=0pt, anchor=center] at (0.30,1.145) {\includegraphics[width=7.4mm]{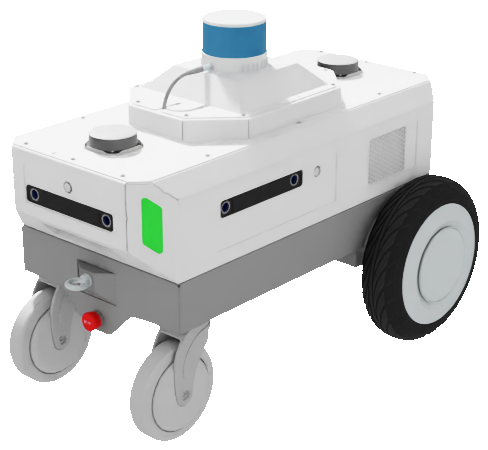}};
  \node[hxseqbox, minimum width=0.72cm, minimum height=3.80cm, anchor=north west] (seqbox) at (4.35,{\ctop+0.07}) {};
  \coordinate (gLm) at (4.71,3.79); \coordinate (gCm) at (4.71,2.715); \coordinate (gOm) at (4.71,1.145);
  \hxGroupN{gL}{(gLm)}{hrBlue}{2}
  \hxGroupN{gC}{(gCm)}{hrGreen}{3}
  \hxGroupN{gO}{(gOm)}{hrOrange}{3}
  \draw[hxflow] (chip.east |- gLm) -- (seqbox.west |- gLm);
  \draw[hxflow] (cctv.east |- gCm) -- (seqbox.west |- gCm);
  \draw[hxflow] (ob.east |- gOm) -- (seqbox.west |- gOm);
  \node[hxmodel, minimum width=1.55cm, minimum height=3.66cm, anchor=north west] (model) at (5.67,\ctop) {\method};
  \draw[hxflowthick] (seqbox.east |- model.west) -- (model.west);
  \node[inner sep=0pt, anchor=north east] (scene) at (17.30,\ctop)
    {\edef\hxtmp{\noexpand\heroFrameC{black!45}{9.02cm}{\heroSceneTrim}{\heroSceneFile}}\hxtmp};
  \draw[hxflowthick] (model.east) -- (scene.west |- model.east);
  \node[font=\sffamily\fontsize{5.5}{6}\selectfont, text=black!55, anchor=south, inner sep=0pt]
    at ($(model.east)!0.5!(scene.west |- model.east)+(0,0.09)$) {waypoints};
  \heroRel{scene}{%
    \heroSceneOverlay
  }
  \node[inner sep=0pt, anchor=south east] (obo) at ($(scene.south east)+(-0.09,0.09)$)
    {\edef\hxtmp{\noexpand\heroFrameC{hrOrange}{2.95cm}{\heroObTrim}{\heroObPathFile}}\hxtmp};
  \heroRel{obo}{\heroTagTR{on-board}}
\end{tikzpicture}%
  \end{minipage}%
    
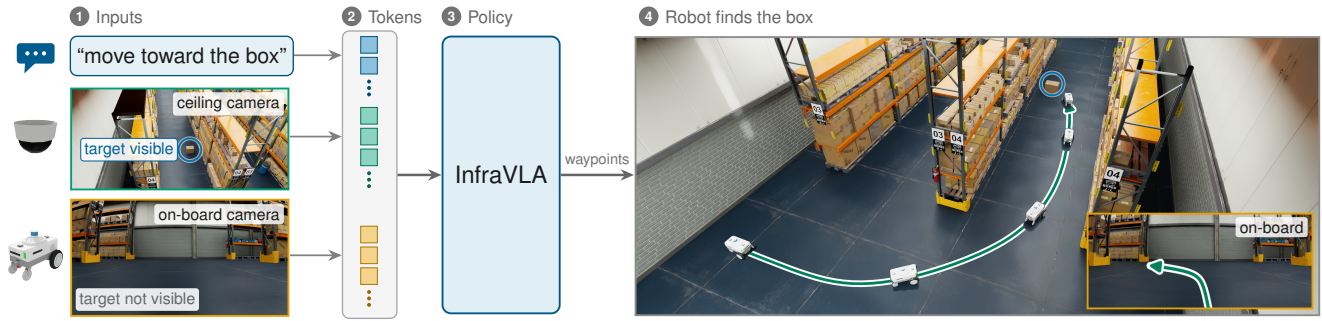
\captionof{figure}{Visualization of our method. The robot is instructed to move toward the box, which is not visible from the start position (left side, bottom image), but from one CCTV camera (marked with a blue ring). The instruction, the ceiling views and the on-board view become one token sequence, from which the policy predicts waypoints. Right: the run seen from above and the same path in the robot's own view at the start (bottom right).}%
    \label{fig:hero}%
  \vspace{1.0mm}%
}

\maketitle

\begin{abstract}
Many indoor environments in which robots operate, such as warehouses, offices, and hospitals, already have cameras installed. They observe parts of the building that the robot cannot see from where it stands, yet navigation policies, including recent vision-language-action (VLA) models, do not use them. We propose \method, an end-to-end method that adapts a pretrained navigation VLA to such static infrastructure views: a closed-circuit television (CCTV) encoder turns each external view into tokens of the input sequence. Because the views matter only at rare decision points, fine-tuning alone did not make the policy use them in our experiments; we therefore train in two stages, on demonstrations with upsampled counterfactual data and then on recovery data. We evaluate on two simulated warehouse tasks, finding an object named in the instruction and rerouting around blocked aisles, where the deciding information is often visible only to the infrastructure cameras. Tested in distribution, \method reached a success rate of 100\% on both, against 34.0\% and 73.6\% for a baseline without CCTV input. On out-of-distribution test sets it reached 88.2\% and 88.9\%. On a real quadruped fine-tuned with under 10 minutes of demonstrations, the policy reached 83.3\% against 29.2\% for the on-board-only baseline.

\end{abstract}

\section{Introduction}
\label{sec:introduction}

Warehouses, offices, hospitals, train stations, and construction sites are commonly equipped with \ac{cctv} cameras. A robot deployed there sees only what lies in front of its own cameras, whereas the building's cameras cover the aisles and rooms around it, in real time. A map built earlier, for example with simultaneous localization and mapping, goes stale in an environment that keeps changing, while the cameras always show the current state. Yet many navigation methods, including recent \ac{vla} models, do not use this information from already existing \ac{cctv} cameras.

Consider an example. A robot in a warehouse is told to go to a box. The box is out of sight, and the person giving the instruction may not know where it is either. With only its on-board cameras, the robot has to search the aisles one by one, and it may enter an aisle whose far end is blocked by other robots or workers and have to backtrack. The ceiling cameras see the box and the blocked aisle. A policy that could access them would drive to the box directly and avoid the blocked aisle. \Cref{fig:hero} shows what this could look like: the robot heads directly to a box that is initially not visible in its on-board view but is visible in the external views.

\acp{vla} are a natural place to add such views. They reuse the pretrained vision and language backbones of \acp{vlm} and are pretrained on large robot datasets~\cite{kimOpenVLAOpenSourceVisionLanguageAction2024a,hiroseOmniVLAOmniModalVisionLanguageAction2025b}: their vision encoders extract semantic and spatial information from an image, their \ac{llm} backbones take in the language instruction, and an action head predicts the actions. A further view can therefore enter the model directly, end to end, rather than through fixed interfaces between separate components, whose errors can compound~\cite{ChenChallengesAndFrontiers24}. It also leaves what the policy takes from that view to training, rather than to a module that decides in advance what to pass on and what to discard. We therefore build on a pretrained navigation \ac{vla}; to our knowledge, none has been extended with static and external views from the infrastructure in an end-to-end manner.

However, only adding a \ac{cctv} encoder to the architecture is not enough. We found that the policy also needs a training recipe that makes it use the encoder and the tokens it generates. In our datasets the data points that require the \ac{cctv} input are rare: the policy only needs it at a few decision points of a trajectory, for example when it has to change direction, and can often predict the next action from the on-board view alone by following the path it is already on. A policy that ignores the \ac{cctv} input can therefore still achieve a low loss on the training data, a form of shortcut learning~\cite{geirhosShortcutLearningDeep2020} related to the copycat problem~\cite{wenFightingCopycatAgents2020}.

In this paper, we propose \method, an end-to-end method for adapting a pretrained navigation \ac{vla} to external camera inputs, together with a training recipe that addressed these failures in our experiments. The method builds on a navigation \ac{vla} pretrained on on-board views alone, here OmniVLA~\cite{hiroseOmniVLAOmniModalVisionLanguageAction2025b}, which maps an on-board image and a goal given as language, an image or a pose to waypoints. A dedicated \ac{cctv} encoder turns each external view into tokens, downsampled so that several cameras fit into the backbone's context, and inserts them next to the on-board and language tokens. We train the policy in two stages. In stage one, we fine-tune on demonstrations that contain counterfactual pairs~\cite{glossopCASTCounterfactualLabels2025}, e.g., two trajectories from the same spawn point that differ only in the language instruction or in the configuration of the scene, and we upsample the counterfactual snippets in which the two paths split. In stage two, we continue training on recovery data mined within those snippets: we roll out the stage-one policy from a position inside a snippet and, once it derails, a supervisor returns it to the demonstrated path.

We validate the method in simulation and on a real robot. In a semantic task, in which the policy has to navigate to an object named in the instruction, \method achieved a \ac{sr} of 100\%, compared with 34.0\% for a baseline without \ac{cctv} input. In a spatial task, in which blocked aisles force the robot to reroute, it achieved 100\% against 73.6\% for the baseline, and took the shortest route more often. Policies trained with configurations held out reached 88.2\% on unseen object placements and 88.9\% on unseen combinations of blocked aisles. On a Boston Dynamics Spot fine-tuned with less than 10 minutes of demonstrations, \method reached 20 of 24 targets, compared with 7 of 24 for the on-board-only baseline.

To summarize, our contributions are:
\begin{itemize}
\item A \textbf{model architecture} that extends a pretrained navigation \ac{vla} to a multi-view setting with several static \ac{cctv} cameras, trained end to end.
\item A \textbf{two-stage training recipe} built on upsampled counterfactual data and recovery data, without which the policy did not learn to use the multi-view input in our experiments.
\item A demonstration that the trained policy completes the tasks it was trained for and holds up on held-out configurations, in simulation and in a real-world experiment.
\end{itemize}

\section{Related Work}
\label{sec:related_work}

\textbf{Navigation with infrastructure cameras.} Traditional navigation stacks often use external cameras only to improve one component: localization~\cite{flogelInfrastructureAidedLocalizationState2022a}, \acp{bev} built from roadside cameras~\cite{xuCoBEVTCooperativeBirds2022b}, detection of humans~\cite{bultmannAnticipatingHumanBehavior2025} or other robots~\cite{ravankarIntelligentRobotGuidance2016}, or path planning~\cite{otaEfficientExplorationConstrained2020a}.
In driving, the cameras are relevant by construction, since they observe the vehicle or its road; indoor infrastructure cameras observe regions that may not matter for the current action. Ota et al.~\cite{otaEfficientExplorationConstrained2020a} train a \ac{cnn} on top-down views of the environment to predict the shortest path that the robot then follows, and Robot-Relay~\cite{robinson2026robot} servos a robot with external cameras through a network of camera-observed spaces linked by handover regions. Closest to our setting, SurveilNav~\cite{yuSurveilNavCollaborativeObject2026} selects cameras to build frontier and value maps for object search and plans with a modular pipeline, and Indoor-R2X~\cite{yangIndoorR2XIndoorRobottoEverything2026a} turns camera and \ac{iot} events into logs from which an \ac{llm} planner derives atomic actions, such as ``turn left''. Our method is one end-to-end pipeline in which the policy can use the views for any purpose, and involves the foundation model in the local navigation itself. To our knowledge, none of these methods combines external cameras with an end-to-end trained foundation model that takes language goals indoors.

\textbf{Navigation VLAs and infrastructure-aided driving.} End-to-end policies, in contrast, map observations directly to actions. Navigation policies increasingly build on foundation models, for example by fine-tuning a \ac{vlm} to output atomic instructions~\cite{chengNaVILALeggedRobot2025}. OmniVLA~\cite{hiroseOmniVLAOmniModalVisionLanguageAction2025b} extends OpenVLA~\cite{kimOpenVLAOpenSourceVisionLanguageAction2024a} with goal-image and goal-pose inputs and showed that its 100M-parameter edge version adapts to satellite imagery as a further goal modality. Our \ac{cctv} views are not a goal modality; they add environmental information that lets the policy complete tasks it could not before. The nearest multi-view navigation \ac{vla}, MM-Nav~\cite{xuMMNavMultiViewVLA2026}, uses four cameras, all mounted on the robot, so its observations stay local. In driving, UniV2X~\cite{yuEndtoEndAutonomousDriving2024} fuses roadside cameras into an end-to-end pipeline without foundation models, and V2X-VLM~\cite{youV2XVLMEndtoEndV2X2024} lets a \ac{vlm} predict waypoints from vehicle and infrastructure views; both operate outdoors with cameras that observe the vehicle's own road, and neither adapts a pretrained navigation \ac{vla}.

\textbf{Adding modalities to VLAs.} Our architecture follows the recipe used to inject new modalities into pretrained \acp{vla}, such as point clouds~\cite{liPointVLAInjecting3D2025a}, tactile sensing~\cite{yuForceVLAEnhancingVLA2025}, and event cameras~\cite{zhaiEVLAEventAugmentedVisionLanguageAction2026}: a modality-specific encoder feeds the backbone, which stays frozen or is trained parameter-efficiently, because data with the new modality is scarce compared with on-board image data. Methods differ in where the signal enters, overlaid on the RGB image~\cite{zhaiEVLAEventAugmentedVisionLanguageAction2026}, as tagged tokens in the backbone~\cite{yuForceVLAEnhancingVLA2025}, or directly in the action head~\cite{liPointVLAInjecting3D2025a}. We follow the same pattern, inserting the extra views as tokens in the backbone. The difference is the sensor: these modalities are recorded on the robot; infrastructure cameras are not, so the policy must also learn which view matters when.

\textbf{Shortcut learning and counterfactual data.} We train our policy by behavior cloning, supervised learning on demonstrations of the task, which brings two known failure modes. The first is shortcut learning~\cite{geirhosShortcutLearningDeep2020}: a model exploits incidental correlations that minimize the training loss but often fail in closed-loop rollouts. Copycat behavior~\cite{wenFightingCopycatAgents2020}, in which an agent predicts its next action from its own history, is one instance, and we observed a form of it (\cref{sec:experiments}). Wen et al.~\cite{wenKeyframeFocusedVisualImitation2021} upweight the keyframes at which the demonstrated action changes; we upsample counterfactual snippets, which also provide the alternative outcome. Counterfactual data has been used for vision-and-language navigation~\cite{fuCounterfactualVisionandLanguageNavigation2025}. Closest to our work, CAST~\cite{glossopCASTCounterfactualLabels2025} synthesizes counterfactual instruction-label pairs with a \ac{vlm}, whereas our pairs are two demonstrations from the same start under a different instruction or scene, so no synthetic annotation is needed.

\textbf{Covariate shift and recovery data.} A policy's own rollouts drift into states the supervisor never visited, where errors compound. DART~\cite{laskeyDARTNoiseInjection2017a} injects noise into the demonstrations, so that they cover the states a drifting policy visits. DAgger-family methods~\cite{rossReductionImitationLearning2011a} relabel states visited by earlier policies, and HG-DAgger~\cite{kellyHGDAggerInteractiveImitation2019} lets the supervisor take over at a safety threshold. Our recovery data follows HG-DAgger with a distance-to-path threshold, is mined at the counterfactual snippets, and is labeled by an \astar oracle.

\section{Method}
\label{sec:method}

Our approach is to extend a pretrained navigation \ac{vla} with an encoder for the infrastructure views (\cref{sec:method-architecture}) and to fine-tune it on demonstrations that include those views. The fine-tuning follows a two-stage recipe (\cref{sec:method-stage1,sec:method-stage2}) that enables the model to use the external views. We first state the problem (\cref{sec:method-problem}). 

\subsection{Problem Setting}
\label{sec:method-problem}
A robot operates in a single fixed facility $E$ equipped with $N$ static cameras. The intrinsics and extrinsics of the cameras are fixed but not provided to the policy. How an infrastructure view relates to the robot's frame has to be inferred from data. A scenario $\xi = (c, \mathcal{G}, s_0)$ consists of a configuration $c$, the mutable part of the facility such as the placement of obstacles or goal objects, a task specification $\mathcal{G}$ given as a goal pose, a goal image, a language instruction, or a subset of these, and the robot's initial state $s_0$. At each decision step $t$ the policy receives the observation $o_t = (I^{o}_t, \{I^{c_i}_t\}_{i=1}^{N})$, the on-board image and the $N$ infrastructure images, and predicts an action $a_t = \pi(o_t, \mathcal{G})$, a sequence of waypoints that a low-level controller executes. A trajectory succeeds if the robot comes within a success radius $\varepsilon_r$ of the goal coordinates (\cref{tab:settings}), and the objective is the policy that maximizes the expected success over the facility's scenario distribution. The facility is fixed by design: we fine-tune a pretrained policy for deployment in one facility, which keeps the required data at an amount a human can collect in a short time.

\subsection{Architecture}
\label{sec:method-architecture}

\begin{figure}[t]
  \centering
  \resizebox{0.88\columnwidth}{!}{\input{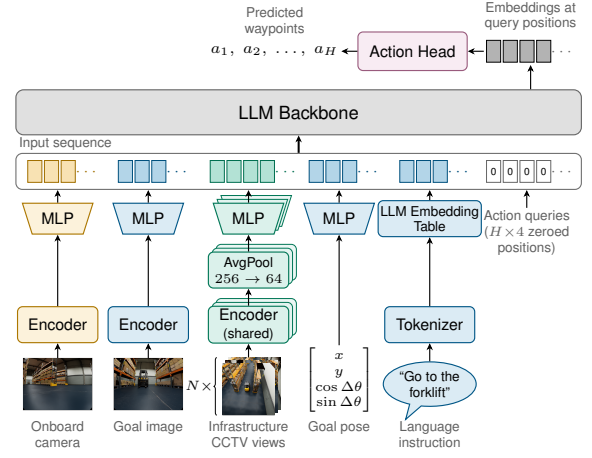}}
  \caption{Architecture. Green: our extension, the infrastructure images, encoded per view, pooled from 256 to 64 tokens, and inserted after the on-board tokens. Orange: the on-board image pipeline; blue: the task specification; both as in OmniVLA~\cite{hiroseOmniVLAOmniModalVisionLanguageAction2025b}.
  \vspace{-18pt}
}
  \label{fig:arch}
\end{figure}

\textbf{Base model.} We build on OmniVLA~\cite{hiroseOmniVLAOmniModalVisionLanguageAction2025b}, a navigation \ac{vla} derived from OpenVLA~\cite{kimOpenVLAOpenSourceVisionLanguageAction2024a}. The on-board image is encoded by pretrained DINOv2~\cite{oquabDINOv2LearningRobust2024a} and SigLIP~\cite{zhaiSigmoidLossLanguage2023} encoders, whose concatenated features are projected into the embedding space of a LLaMA~2 7B backbone~\cite{touvronLLaMAOpenEfficient2023}. A goal image uses the same encoder, a 2D goal pose is projected by a \ac{mlp}, and the language instruction enters through the LLaMA tokenizer. The sequence ends with $4H$ zero tokens: the model predicts $H = 8$ future waypoints, and each waypoint is a quadruple $(x, y, \cos\theta, \sin\theta)$, the offset and the yaw in the robot's frame, so one waypoint requires four tokens. A regression head reads the backbone outputs at these positions and predicts the $4H$ values.

\textbf{Adding infrastructure views.} We define three requirements to design our extension: the cameras are processed end to end as part of the model, the design scales to a reasonable number of cameras, and it is trainable from little data. They lead to three decisions; the resulting architecture is shown in \cref{fig:arch}. First, the \ac{cctv} images are encoded by a dedicated encoder rather than turned into a text or graph description by a separate segmentation or \ac{vlm} module, so that what the policy takes from a view is learned during training, with no intermediate description whose errors can propagate~\cite{ChenChallengesAndFrontiers24}. We expect the tasks of the on-board encoder and the \ac{cctv} encoder to differ. In the on-board view the content that matters, the free space and the obstacles ahead, sits in a predictable part of the image. In a \ac{cctv} view it can be a target object, a blocked aisle, or the robot itself, anywhere in the frame and often small, depending on where the robot is and where it is going, so the encoder has to preserve detail across the whole view. The \ac{cctv} tokens are inserted after the on-board observation and goal-image tokens, keeping those at the positions OmniVLA was pretrained with. Second, the \ac{cctv} encoder's 256 tokens per view are reduced to 64 by $2 \times 2$ average pooling of the spatial feature map, so that several views fit into the backbone's context. Third, the \ac{cctv} encoder and its projector are initialized from the on-board encoder and projector, so that their features are aligned with the backbone from the start and the encoder only has to learn the residual between the two tasks. All linear layers are fine-tuned with LoRA~\cite{huLoRALowRankAdaptation2021}, which Kim et al.~\cite{kimOpenVLAOpenSourceVisionLanguageAction2024a} showed to be effective for adapting \acp{vla}. The rest of the model is unchanged.

\textbf{Loss.} We keep OmniVLA's loss but drop its language-following loss, which in our setting produced artifacts in the model's last predicted waypoint, pulling it toward the origin. What remains is the waypoint regression with a smoothness term,
\begin{equation}
  \mathcal{L} = \frac{1}{H} \sum_{i=1}^{H} \norm{a_i^{\text{ref}} - \hat{a}_i}_2^2
  + \frac{\lambda}{H-1} \sum_{i=1}^{H-1} \norm{\hat{a}_{i+1} - \hat{a}_i}_2^2,
  \label{eq:loss}
\end{equation}
where $\hat{a}_i$ is the $i$-th predicted waypoint, $a_i^{\text{ref}}$ the supervisor's, and $\lambda = 0.1$ scales the auxiliary smoothness term.

\subsection{Stage 1: Counterfactual Snippets and Upsampling}
\label{sec:method-stage1}
We want the policy to make use of the \ac{cctv} views, so training has to give it an incentive to do so. Under behavior cloning, the incentive has to come through the training loss. The policy minimizes the expected negative log-likelihood $\mathcal{R}(\pi) = \E_{(o,a) \sim \mathcal{D}_{\text{E}}}[-\log \pi(a \mid o)]$ over the supervisor's data $\mathcal{D}_{\text{E}}$, and the best achievable risk is the conditional entropy $\mathcal{H}(A \mid O)$ of the action given the observation. Following CAST~\cite{glossopCASTCounterfactualLabels2025}, we split the observation into $O = (X, Z)$, where $Z$ is the \ac{cctv} images, together with the language instruction when the goal is an object named in it, and $X$ is the rest of the observation, e.g., the on-board image. The best policy that ignores $Z$ reaches $\mathcal{H}(A \mid X)$, the best policy that uses it $\mathcal{H}(A \mid X, Z)$, and their difference is the conditional mutual information between the action and $Z$ given $X$,
\begin{equation}
  \mathcal{H}(A \mid X) - \mathcal{H}(A \mid X, Z) = I(A; Z \mid X).
  \label{eq:mi}
\end{equation}
This is the incentive: $I(A; Z \mid X)$ is the reduction in uncertainty about the action that $Z$ brings beyond $X$, and thus the loss a policy can save by using $Z$. If it is zero, ignoring the infrastructure cameras is optimal for the training objective. Counterfactual data~\cite{glossopCASTCounterfactualLabels2025} raises $I(A; Z \mid X)$, and is how we create the incentive.

\textbf{Pairs and snippets.} A counterfactual pair is two samples $\{(x, z_1, a_1), (x, z_2, a_2)\}$ with $z_1 \neq z_2$ and $a_1 \neq a_2$: they share $X$, so only $Z$ explains the correct action. We generate such pairs directly. In simulation, the robot starts from the same spawn point, and either the language instruction or the configuration $c$ is changed before the supervisor's next trajectory is recorded. The two trajectories are identical until the path splits, so the on-board view ($X$) is the same up to that point and only $Z$, the \ac{cctv} views or the changed instruction, explains the different action. On the real robot we replicate this by starting demonstrations from the same position. Where the same start cannot be reproduced, the pairs can instead be synthesized, as CAST~\cite{glossopCASTCounterfactualLabels2025} does. We call the part of a trajectory around the split a counterfactual snippet. Given a pair of trajectories, we compare the two robot poses at the same time step and, where their heading or their distance differs by more than a threshold, extract a window of fixed length around that step. Metric, threshold, and window size are experiment-specific (\cref{tab:settings}).

\textbf{Upsampling.} The snippets are the part of the data where $Z$ matters, and they are a small fraction of it: elsewhere the next action can usually be derived from the on-board view alone, for example by continuing straight along an aisle. $Z$ therefore matters rarely for the loss, so a policy trained without upsampling reaches a low loss without using it, yet it matters decisively for success, because the choice made at the split decides whether the robot reaches the goal. Following Wen et al.~\cite{wenKeyframeFocusedVisualImitation2021}, who upweight the keyframes, we upsample the snippets in the data loader so that they make up around half of the sampling mass, which widens the gap in achievable loss between a policy that uses $Z$ and one that does not and so adds optimization pressure to use it. Stage 1 fine-tunes OmniVLA on the complete demonstrations of the task, with the counterfactual snippets they contain upsampled as described, and yields the policy $\pi_1$.

\subsection{Stage 2: Recovery Data}
\label{sec:method-stage2}
Behavior cloning assumes i.i.d.\ samples, but during a rollout small prediction or execution errors can bring the robot into states the demonstrations do not cover, a covariate shift under which errors compound. We counter it with recovery data in the style of HG-DAgger~\cite{kellyHGDAggerInteractiveImitation2019}: the stage-1 policy is rolled out, a supervisor takes over once the rollout leaves the demonstrated path, and the supervisor's way back is added to the training data. We mine recovery data only within the counterfactual snippets. OmniVLA is pretrained on plenty of cruising, whereas the counterfactual snippets, where the policy has to act on what the infrastructure cameras show, are the part of our data that is new to it, so we expect compounding errors there.

\textbf{Mining rule.} For each snippet, $\pi_1$ is spawned at a pose inside the snippet and rolled out, or at the trajectory's original start if it does not derail from there. The rollout derails at the first step $t_{\text{off}}$ at which its distance to the demonstrated trajectory exceeds a predefined distance $d_{\text{r}}$; $t_{\text{on}}$ is the last earlier step within a smaller predefined distance $\epsilon_{\text{on}}$ of it. The supervisor takes over at a step drawn uniformly between $t_{\text{on}}$ and $t_{\text{off}}$ and drives back until the robot is within $\epsilon_{\text{on}}$ of the demonstrated trajectory; only this return segment is kept for training. Mining stops after a set number $K$ of recovery trajectories. The supervisor is the \astar oracle in simulation; on a real robot it would be the human operator. Stage 2 continues training $\pi_1$ on a mixture of recovery data and counterfactual snippets and yields $\pi_2$.

\section{Experiments}
\label{sec:experiments}

\begin{table}[t]
  \caption{Hyperparameters of the three experiments: data, training, recovery mining, snippet detection, and success criterion.}
  \label{tab:settings}
  \centering\scriptsize
  \setlength{\tabcolsep}{1.5pt}
  \begin{tabular}{@{}lccc@{}}
    \toprule
     & Semantic (sim) & Spatial (sim) & Semantic (real) \\
    \midrule
    CCTV cameras & 4 & 4 & 2 \\
    Goal modality & language & 2D pose & language \\
    Success & \qty{1}{\metre} to object & \qty{1}{\metre} to goal & human-judged \\
    Test rollouts & 50 & 72 & 24 \\
    \midrule
    Trajectories (train / val.) & 552 / 48 & 481 / 46 & 75 (all) \\
    Training data (h) & 3.03 & 6.89 & 0.16 \\
    CF snippet share & 5.7\% & 10.2\% & 11.8\% \\
    CF mining rule & heading $> 1^\circ$ & dist.\ $> \qty{0.75}{\metre}$ & heading $> 8^\circ$ \\
    Snippet window (before / after) & 3 / 0 & 16 / 8 & 4 / 0 \\
    \midrule
    CF share, upsampled & 50\% & 47\% & 34.9\% \\
    Steps stage 1 $\rightarrow$ 2 & 15k $\rightarrow$ 5k & 15k $\rightarrow$ 5k & 3k \\
    Recovery traj.\ $K$ & 48 & 6--48 & - \\
    Stage-2 mix (CF/other/recovery, \%) & 80/0/20 & 50/0/50 & - \\
    \bottomrule
  \end{tabular}
  \vspace{-15pt}
\end{table}

We evaluate the method in two simulated tasks, reaching an object named in the instruction and rerouting around blocked aisles, both also tested on out-of-distribution data, and validate the findings in a real-robot experiment.

\subsection{Setup}
\label{sec:exp-setup}
\textbf{Simulation.} The two simulated tasks run in an Isaac Sim warehouse of \qty{50.8}{\metre} by \qty{14.1}{\metre} (\cref{fig:semantic}, left): three open areas, one at each end and one in the middle, separated by two rows of three shelves; each row forms two aisles, \qty{16.2}{\metre} long and \qty{3.9}{\metre} wide. Four ceiling cameras in the corners, \qty{7.89}{\metre} high, render $720 \times 720$ images. We use a Nova Carter, a robot with a differential drive at up to \qty{1}{\metre\per\second}. The supervisor plans with \astar on a \qty{0.1}{\metre} grid, similar to Cai et al.~\cite{caiNavDPLearningSimtoReal2025a}, with obstacles inflated by the robot radius to avoid collisions, an edge cost from a \ac{edf} that keeps the path away from them, and natural cubic splines for a smooth path, and follows the path with a pure-pursuit controller at \qty{30}{\hertz}; data is recorded at \qty{3}{\hertz}.

\textbf{Execution.} At test time the policy runs at \qty{3}{\hertz}; following OmniVLA, its fifth predicted waypoint is turned into a forward and an angular velocity with the inverse unicycle model and passed to the robot's velocity controller.

\textbf{Training and evaluation.} All policies start from the OmniVLA checkpoint at 120k steps and are fine-tuned with LoRA (rank 16, $\alpha = 8$) on all linear layers, using AdamW with learning rate $10^{-4}$, batch size 30, $\lambda = 0.1$, and $H = 8$; recovery data in simulation is mined with $d_{\text{r}} = \qty{0.5}{\metre}$ and $\epsilon_{\text{on}} = \qty{0.05}{\metre}$. The experiment-specific settings are in \cref{tab:settings}. Every result is one training run, and all success rates come from closed-loop rollouts, in which the policy drives the robot and its observations depend on its own earlier actions.

\textbf{Baseline.} The baseline in each experiment is the same model fine-tuned on the same data with upsampling. The baseline has no \ac{cctv} encoder and no \ac{cctv} input, so that the comparison isolates the effect of the added views.

\subsection{Semantic Task: Reaching a Named Object}
\label{sec:exp-semantic}
\begin{figure}[t]
  \centering
  \renewcommand{\arraystretch}{0}%
  \newcommand{\cv}[2]{\includegraphics[height=60.7pt,trim=#2,clip]{cctv_lang_0508tA_#1}}%
  \includegraphics[height=122.6pt,trim=5.58 4.84 5.59 4.84,clip]{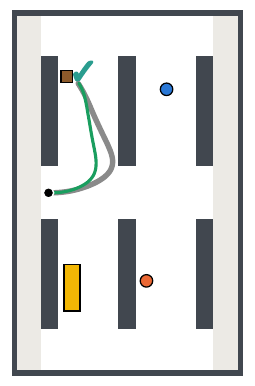}\hspace{5pt}%
  \begin{tabular}[b]{@{}c@{\hspace{1.2pt}}c@{}}
    \cv{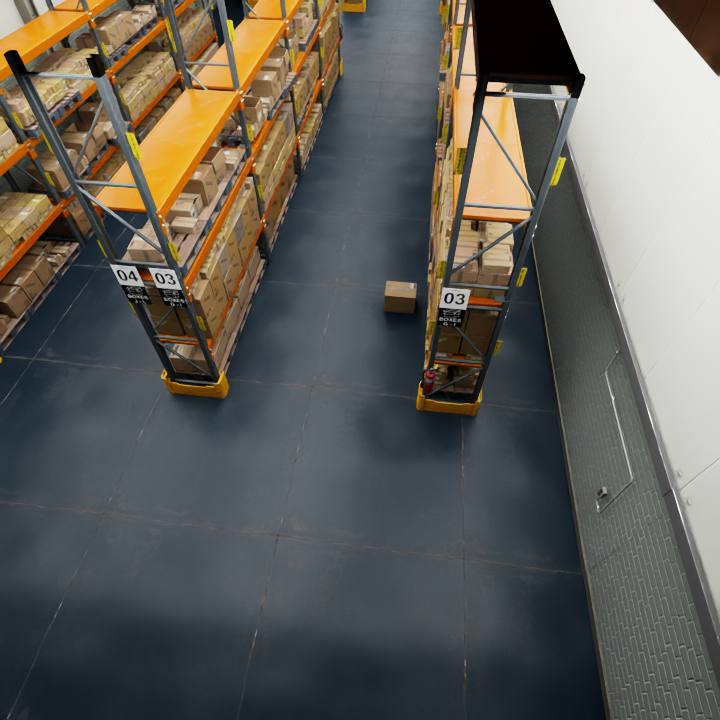}{0 288 144 0} & \cv{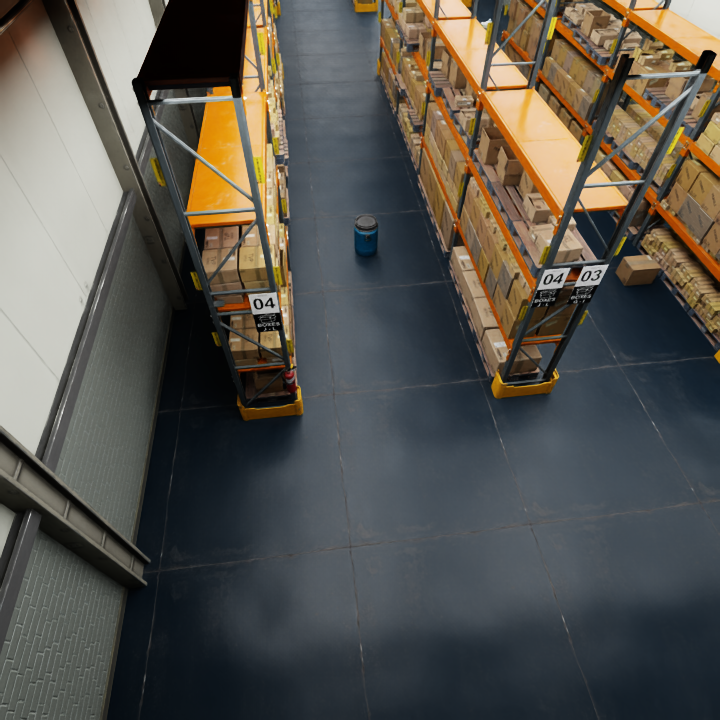}{144 288 0 0}\\[1.2pt]
    \cv{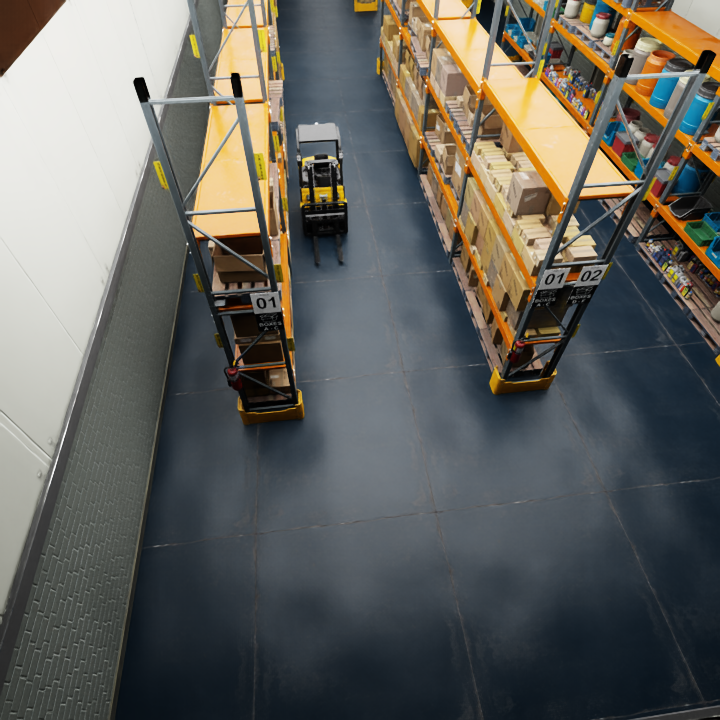}{144 288 0 0} & \cv{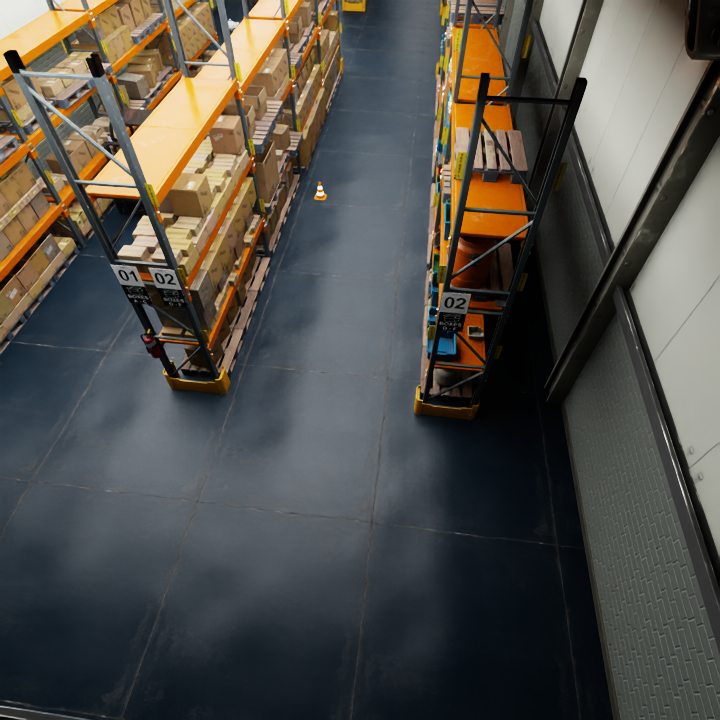}{0 288 144 0}
  \end{tabular}
  \caption{Semantic task. Left: two rollouts of one scene in plan view, both instructed to reach the box (brown square) and both reaching it. From the black start dot, \method after stage 1 (gray) passes close to a shelf (dark gray) and after stage 2 (green) keeps its distance. Right: what the four ceiling cameras see in that same scene at the start; the box is top left.}
  \label{fig:semantic}
\end{figure}

This experiment tests whether the policy can follow a language instruction with what the infrastructure cameras show: it has to recognize the named object in the \ac{cctv} views and drive to it. Four objects (a barrel, a forklift, a box, and a traffic cone) are placed one per aisle, in the outer half of the aisle, where only the ceiling cameras see them (\cref{fig:semantic}). Every scene draws a new assignment of the four objects to the four aisles and, for each object, a position at random within its half. The robot starts at a fixed pose in the middle area, facing east, and receives ``move toward the \{object\}''. Every scene yields four demonstrations, one per object, which form the counterfactual pairs. The test set, 13 such scenes with four instructions each, is generated with the same procedure but repeats no training scene; two rollouts without an \astar solution were discarded, leaving 50.
\begin{table}[t]
  \caption{Semantic task, 50 rollouts. SR: success within \qty{1}{\metre} of the object. Clearance: mean over rollouts of the closest distance to a shelf (demonstrations: \qty{1.47}{\metre}).}
  \label{tab:semantic}
  \centering\footnotesize
  \begin{tabular}{@{}lcc@{}}
    \toprule
    Policy & SR (\%) $\uparrow$ & Clearance (m) $\uparrow$ \\
    \midrule
    Fine-tuned OmniVLA~\cite{hiroseOmniVLAOmniModalVisionLanguageAction2025b} (no CCTV) & 34.0 & 0.67 \\
    \method, no upsampling        & 36.0 & 0.62 \\
    \method, upsampling (stage 1) & \textbf{100.0} & 0.71 \\
    \quad $+$ recovery (stage 2) & \textbf{100.0} & \textbf{1.27} \\
    \bottomrule
  \end{tabular}
  \vspace{-5pt}
\end{table}

\begin{table}[!t]
  \caption{Semantic task, unseen instruction words: SR over 12 rollouts per word, for two synonyms of each training noun.}
  \label{tab:synonyms}
  \centering\footnotesize
  \begin{tabular}{@{}lcccc@{}}
    \toprule
    Object & Synonym 1 & SR (\%) & Synonym 2 & SR (\%) \\
    \midrule
    Forklift     & lift truck  & 91.7  & vehicle     & 66.7 \\
    Barrel       & canister    & 91.7  & oil drum    & 75.0 \\
    Box          & package     & 91.7  & crate       & 83.3 \\
    Traffic cone & safety cone & 100.0 & witch's hat & 25.0 \\
    \bottomrule
  \end{tabular}
  \vspace{-15pt}
\end{table}

In \cref{tab:semantic}, the baseline without \ac{cctv} input reaches an \ac{sr} of 34.0\%, only 9 percentage points above the 25\% of guessing an aisle: without the infrastructure views the task is barely better than a guess. Adding the \ac{cctv} encoder alone does not help: without upsampling the policy reaches 36.0\%, and its validation loss on the counterfactual snippets is 8.9 times its overall validation loss, so it has not learned the decision points. With the snippets upsampled to half the sampling mass, the snippet loss falls to 0.3 times the overall loss, and the stage-1 policy reaches an \ac{sr} of 100\%. The stage-2 policy keeps the \ac{sr} and raises the mean clearance to the shelves (\cref{fig:semantic}) from \qty{0.71}{\metre} to \qty{1.27}{\metre} (demonstrations: \qty{1.47}{\metre}): the recovery data corrects small deviations before they compound, so the policy stays closer to the demonstrated path and keeps a safer distance from the shelves. The \ac{cctv} encoder raises inference time on an RTX PRO 6000 from 95 to \qty{190}{\milli\second} with four \ac{cctv} views.

\textbf{Out-of-distribution tests.} We tested generalization to configurations never seen in training by retraining the policy with every scene that has the forklift in the top-left aisle of \cref{fig:semantic} held out, six of the 24 object-to-aisle assignments. On these scenes it still reaches 99.0\% when sent to one of the other three objects, each in an aisle it had been seen in during training, and 88.2\% when sent to the forklift in the aisle it had never been seen in (\cref{tab:gen-semantic}).

\begin{table}[!t]
  \caption{Semantic task, out-of-distribution tests. A separate policy trained through stage 2 with every scene that puts the forklift in the top-left aisle held out.}
  \label{tab:gen-semantic}
  \centering\footnotesize
  \setlength{\tabcolsep}{4pt}
  \begin{tabular}{@{}lcc@{}}
    \toprule
    Evaluation set & Rollouts & SR (\%) $\uparrow$ \\
    \midrule
    In-distribution              & 50  & 100.0 \\
    Held-out scene, other target & 103 & 99.0 \\
    Held-out scene, forklift     & 34  & 88.2 \\
    \bottomrule
  \end{tabular}
\end{table}

\textbf{Instructions with unseen words.} In this experiment we replaced the object noun by two synonyms per object that never occur in the training data, with 12 rollouts per word, three per aisle (\cref{tab:synonyms}). Seven of the eight synonyms reach an \ac{sr} of 66.7\% or more, against 25\% for guessing an aisle. The language embeddings are thus aligned with the embeddings the \ac{cctv} encoder produces. Together, the tests show that the policy finds and reaches an object its own camera cannot see by recognizing it in the \ac{cctv} views.

\subsection{Spatial Task: Rerouting Around Blocked Aisles}
\label{sec:exp-spatial}
\begin{figure}[t]
  \centering
  \renewcommand{\arraystretch}{0}%
  \newcommand{\cv}[2]{\includegraphics[height=60.7pt,trim=#2,clip]{cctv_spatial_sLgRcfF_#1}}%
  \includegraphics[height=122.6pt,trim=5.58 4.84 5.59 4.84,clip]{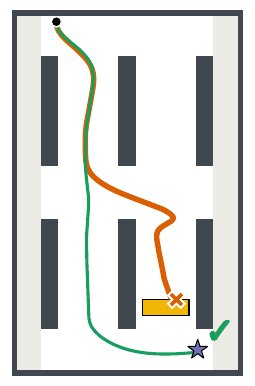}\hspace{5pt}%
  \begin{tabular}[b]{@{}c@{\hspace{1.2pt}}c@{}}
    \cv{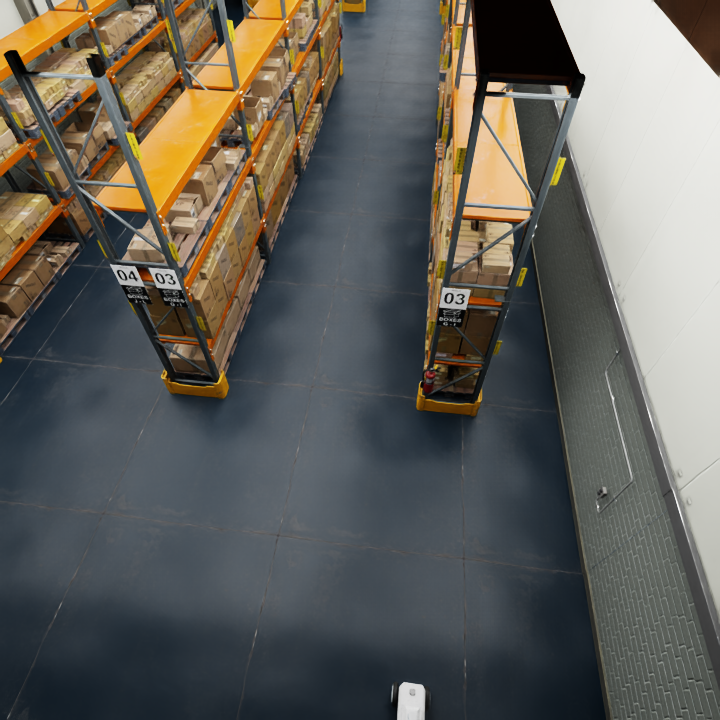}{0 288 144 0} & \cv{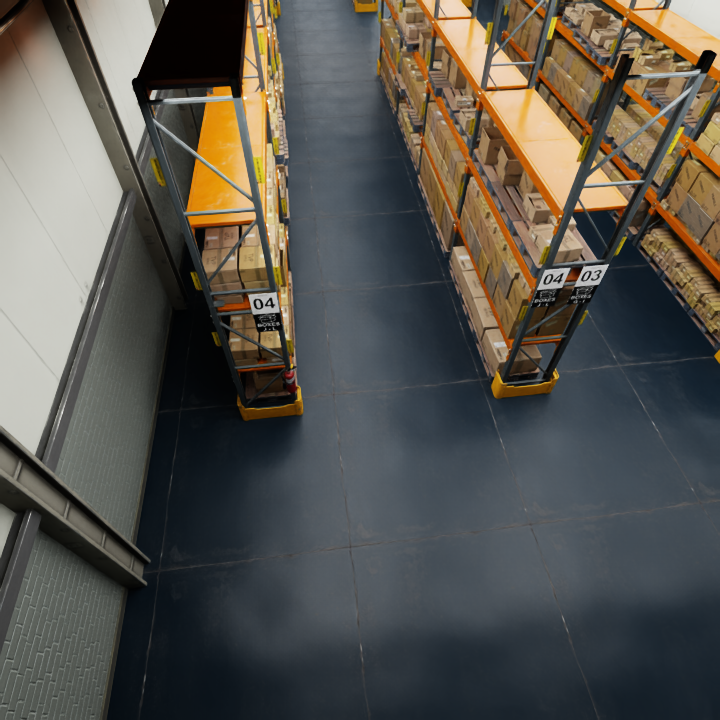}{144 288 0 0}\\[1.2pt]
    \cv{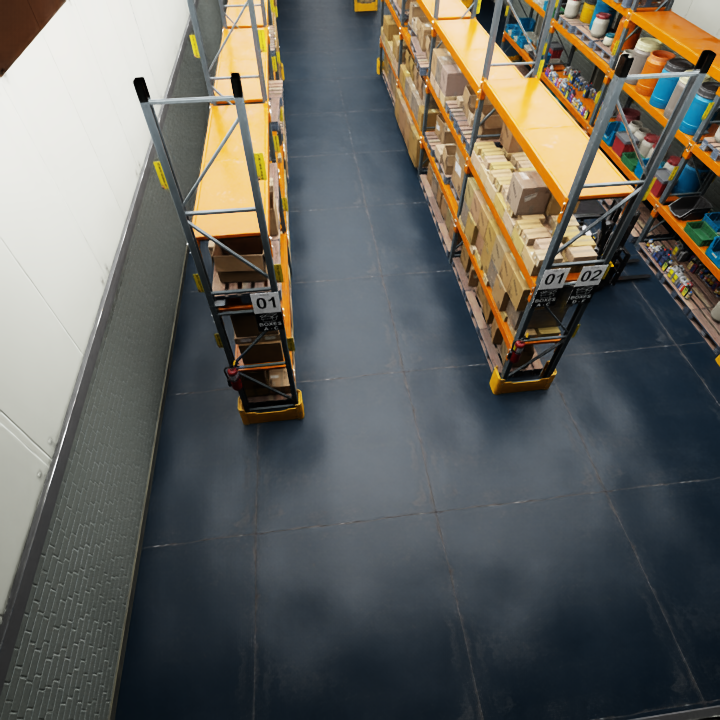}{144 288 0 0} & \cv{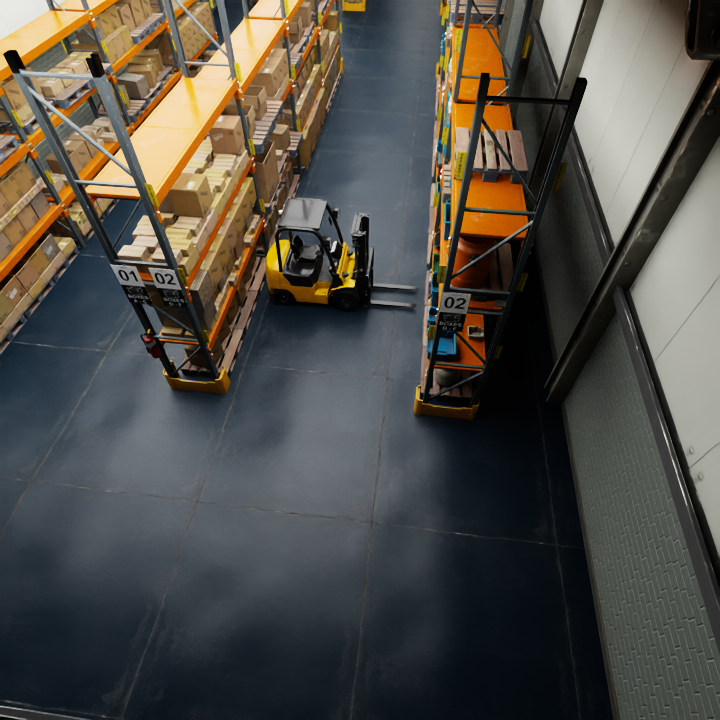}{0 288 144 0}
  \end{tabular}
  \caption{Spatial task. Left: one test scenario in plan view. From the black start dot both policies drive the same route to the middle area; the on-board-only baseline (orange) then enters the aisle sealed by a forklift (yellow), while \method (green) takes the open aisle and reaches the goal (star). Right: the four ceiling views of the same scenario at the start; the forklift that seals the aisle is visible in the bottom-right view.
  \vspace{-20pt}
  }
  \label{fig:spatial}
\end{figure}

This experiment tests spatial understanding: whether the policy reads from the \ac{cctv} views which aisles are blocked by forklifts and reroutes the robot. The robot starts at a pose sampled at random in one end area and receives a 2D goal pose sampled at random in the opposite end area (\cref{fig:spatial}). Up to two forklifts block aisles completely. Each stands at a position drawn at random within the far third of its aisle, where the on-board camera does not see it early; nine blocking patterns exist (none, one of the four aisles, or one aisle in each pair), and the goal is always reachable. Each demonstration is paired with a second one from the same start and goal under a different blocking pattern, so that the forklift positions change the \astar route, which guarantees a counterfactual snippet. We report two metrics on a systematic test set of 72 scenarios (eight blocking patterns, three starts, three goals), absent from the training data: the \ac{sr}, and the share of rollouts that pass through the same aisles as \astar, i.e., take the shortest route.
\begin{table}[t]
  \caption{Spatial task, 72 systematic scenarios. Route: share of rollouts through the same aisles as \astar. $+$ rows: the stage-1 policy trained 5k more steps on the data named (stage 2). CF: counterfactual.}
  \label{tab:spatial}
  \centering\footnotesize
  \begin{tabular}{@{}lcc@{}}
    \toprule
    Policy & SR (\%) $\uparrow$ & Route (\%) $\uparrow$ \\
    \midrule
    Fine-tuned OmniVLA~\cite{hiroseOmniVLAOmniModalVisionLanguageAction2025b} (no CCTV) & 73.6 & 55.6 \\
    \method, no upsampling                  & 76.4 & 51.4 \\
    \method, upsampling (stage 1)           & 77.8 & 44.4 \\
    \quad $+$ CF snippets only, no recovery & 75.0 & 52.8 \\
    \quad $+$ recovery (stage 2), $K = 6$   & \textbf{100.0} & 91.7 \\
    \quad $+$ recovery (stage 2), $K = 12$  & \textbf{100.0} & 93.1 \\
    \quad $+$ recovery (stage 2), $K = 24$  & \textbf{100.0} & \textbf{95.8} \\
    \quad $+$ recovery (stage 2), $K = 48$  & \textbf{100.0} & 94.4 \\
    \bottomrule
  \end{tabular}
  \vspace{-5pt}
\end{table}

\textbf{Results.} \Cref{tab:spatial} shows the baseline, which can solve many scenarios from the on-board view alone, at an \ac{sr} of 73.6\% and 55.6\% shortest routes. With \ac{cctv} input, the stage-1 policy reaches 77.8\% but takes the shortest route in only 44.4\% of the rollouts, and 5k further steps on the counterfactual snippets alone do not solve this. Its typical failure is to leave an aisle into the middle area and cross to the opposite aisle although the aisle ahead is free. The cause is copycat behavior~\cite{wenFightingCopycatAgents2020}: the policy inferred whether to cross from its own heading in the on-board image rather than from the \ac{cctv} views, so that a slight initial turn locked it into crossing. The deterministic \astar demonstrations make this shortcut easy to learn, because every crossing starts with the same slight turn at the same spot. We confirmed it offline: given the on-board image of one rollout of a pair and the \ac{cctv} views of the other, the policy's predictions followed the on-board heading, not the \ac{cctv} views.

\textbf{Recovery data removes the shortcut.} Stage 2 (\cref{sec:method-stage2}) continues training on half recovery data, mined from the policy's own rollouts started inside the snippets, and half counterfactual snippets. The noise of these rollouts breaks the link between a slight heading change and crossing, so the heading no longer predicts the route and the policy has to rely on the \ac{cctv} views. Six recovery trajectories raise the \ac{sr} to 100\% and the shortest-route share to 91.7\%; $K = 24$ gives 95.8\%, against 55.6\% for the baseline and 44.4\% after stage 1. With recovery data the policy not only reaches every goal but takes the \astar route, and both require reading the blocked aisles from the \ac{cctv} views.

\textbf{Out-of-distribution test.} As in \cref{sec:exp-semantic}, we retrained the policy with two of the nine blocking patterns held out, one with a single aisle blocked and one with an aisle blocked in each row, and tested it on the 18 scenarios with a held-out pattern and the 54 with a seen one (\cref{tab:gen-spatial}). After stage 1 the policy succeeds in 33.3\% of the held-out scenarios; after stage 2, with $K = 30$ recovery trajectories mined from the seen patterns only, in 88.9\%, and on the seen scenarios success rises from 75.9\% to 100\%. The recipe that removes the shortcut also holds on blocking patterns the policy never saw. In sum, the policy reads from the infrastructure views which aisles are blocked before its own camera can see the obstacle and takes the open aisle, which can make robot navigation in such indoor environments more effective.

\begin{table}[!t]
  \caption{Spatial task, out-of-distribution test. A separate policy trained with two of the nine blocking patterns held out; recovery data from the seen patterns only. SR on the 54 seen and 18 held-out scenarios.}
  \label{tab:gen-spatial}
  \centering\footnotesize
  \begin{tabular}{@{}lcc@{}}
    \toprule
    Policy & Seen (\%) $\uparrow$ & Held-out (\%) $\uparrow$ \\
    \midrule
    \method, stage 1                       & 75.9 & 33.3 \\
    \quad $+$ recovery (stage 2), $K = 30$ & \textbf{100.0} & \textbf{88.9} \\
    \bottomrule
  \end{tabular}
  \vspace{-15pt}
\end{table}
\subsection{Real Robot}
\label{sec:exp-real}

\begin{figure}[t]
  \centering
  \input{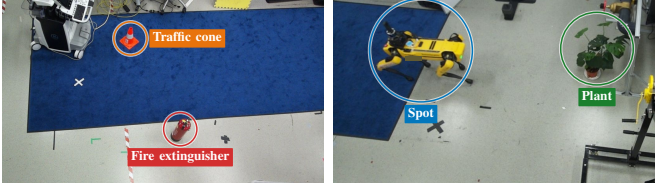}%
  \realSetupCamOne[0.492\columnwidth]\hfill
  \realSetupCamTwo[0.492\columnwidth]
  \caption{The real setup, seen from the two fixed cameras at one instant of a demonstration. Left: \acs{cctv}~1 shows the traffic cone and the fire extinguisher. Right: \acs{cctv}~2 shows the other half of the area, with the robot and the plant. A logo on the robot has been masked for anonymity. 
  }
  \vspace{-5pt}
  \label{fig:real-setup}
\end{figure}

This experiment validates the simulation findings on a real robot. It mirrors the semantic task of \cref{sec:exp-semantic}: the policy has to reach an object named in the instruction that only the infrastructure cameras see.

\textbf{Setup.} A Boston Dynamics Spot is instructed to walk to one of three objects placed in a lab area covered by two fixed cameras, \acs{cctv}~1 and \acs{cctv}~2. The objects are a plant, a fire extinguisher, and a traffic cone (\cref{fig:real-setup}). The cone and extinguisher share their color on purpose. The robot always starts from the same marked pose, which makes counterfactual pairs easy to record and, with a dataset this small, limits the risk of memorizing start poses. A human operator teleoperated the demonstrations: 25 object arrangements with one demonstration per object, 75 trajectories with a total recorded duration of \qty{9}{\minute}\,\qty{53}{\second}. We trained the stage-1 policy and the baseline for 3k steps on all 75; we chose the step count based on the validation loss from a 20/5 split.

\textbf{Test protocol.} Eight new arrangements with three rollouts each were sampled at random under constraints and rebuilt by the operator: objects \qty{1.5}{\metre} to \qty{4.6}{\metre} from the robot and at least \qty{1}{\metre} apart, never visible in the on-board camera at the start, and with a mean object displacement from the nearest training arrangement of at least \qty{1.33}{\metre}, the average over the training set; among the candidates, farthest-point sampling kept the eight most distinct. Whether the robot reached the named object was judged by the operator.

\begin{table}[t]
  \caption{Real robot, 8 test arrangements with one rollout per object (24 rollouts). SR (\%) $\uparrow$ of reaching the named object, judged by the operator, overall and per object.}
  \label{tab:real}
  \centering\footnotesize
  \begin{tabular}{@{}lc@{\hspace{1.1em}}ccc@{}}
    \toprule
    Policy & Overall & Plant & Fire ext.\ & Cone \\
    \midrule
    Fine-tuned OmniVLA~\cite{hiroseOmniVLAOmniModalVisionLanguageAction2025b} & 29.2 & 12.5 & 50.0 & 25.0 \\
    \method & \textbf{83.3} & \textbf{100.0} & \textbf{62.5} & \textbf{87.5} \\
    \bottomrule
  \end{tabular}
  \vspace{-14pt}
\end{table}

\textbf{Results.} \Cref{tab:real} shows the results. The on-board-only baseline succeeds in 7 of 24 rollouts, and its route barely depends on the instruction: it mostly turned right, and walked straight ahead in five rollouts. Our stage-1 policy succeeds in 20 of 24, reaching the plant in all eight scenes and the cone in seven. The fire extinguisher, reduced to a red disc from above, is the weakest object at five of eight, and its failures end at the cone. Only seven training arrangements separate the two red objects onto different sides while both are out of the front view, which may have limited what the policy could learn about them. We did not run stage 2 on the real robot. Its recovery data has to be mined from the policy's own derailments on the training arrangements, and in the simple lab area failures are wrong decisions at the start rather than drift that compounds along a route, so there are few derailments to mine and they cluster on one object. Recovery data collected from staged starting points rather than the policy's own state distribution hurt performance, so we did not pursue this further.

\definecolor{s6rule}{HTML}{9A9A9A}%
\definecolor{s6mark}{HTML}{1F2A44}%
\definecolor{s6dark}{HTML}{101010}%

\makeatletter
\@ifundefined{scenesixcamw}
  {\newlength{\scenesixcamw}\setlength{\scenesixcamw}{\textwidth}}{}
\makeatother
\providecommand{\scenesixthumbw}{0.212\scenesixcamw}

\providecommand{\scenesixpanel}[2][\linewidth]{%
  \setlength{\fboxsep}{0pt}\setlength{\fboxrule}{0.4pt}%
  \textcolor{s6rule}{\fbox{\includegraphics[width=#1]{#2}}}}

\providecommand{\scenesixcropbottom}{0pt}
\newsavebox{\scenesixbox}
\providecommand{\scenesixannotfull}[3][\linewidth]{%
  \begin{tikzpicture}
    \node[anchor=south west,inner sep=0pt,outer sep=0pt] (im)
      {\scenesixpanel[#1]{#2}};
    \begin{scope}[x={(im.south east)},y={(im.north west)}]
      #3
    \end{scope}
  \end{tikzpicture}}
\providecommand{\scenesixannot}[3][\linewidth]{%
  \sbox{\scenesixbox}{\scenesixannotfull[#1]{#2}{#3}}%
  \begin{tikzpicture}
    \clip (0,0) rectangle
      (\wd\scenesixbox,\dimexpr\ht\scenesixbox-\scenesixcropbottom\relax);
    \begin{pgfinterruptboundingbox}
      \node[anchor=south west,inner sep=0pt,outer sep=0pt]
        at (0,-\scenesixcropbottom) {\usebox{\scenesixbox}};
    \end{pgfinterruptboundingbox}
    \draw[s6rule,line width=0.4pt] (0,0.2pt) -- (\wd\scenesixbox,0.2pt);
  \end{tikzpicture}}

\providecommand{\scenesixthumb}[2]{%
  \setlength{\fboxsep}{0pt}\setlength{\fboxrule}{1.0pt}%
  \fcolorbox{white}{white}{\includegraphics[width=#1,trim=0 112 0 0,clip]{#2}}}

\definecolor{s6maskA}{rgb}{0.282,0.290,0.271}%
\definecolor{s6maskB}{rgb}{0.243,0.267,0.263}%
\providecommand{\scenesixthumbmask}[3]{%
  \setlength{\fboxsep}{0pt}\setlength{\fboxrule}{1.0pt}%
  \fcolorbox{white}{white}{%
    \begin{tikzpicture}
      \node[anchor=south west,inner sep=0pt,outer sep=0pt] (tm)
        {\includegraphics[width=#1,trim=0 112 0 0,clip]{#2}};
      \begin{scope}[x={(tm.south east)},y={(tm.north west)}]
        #3
      \end{scope}
    \end{tikzpicture}}}

\providecommand{\scenesixleader}[2]{%
  \draw[line width=1.7pt,s6dark,opacity=0.40] #1 -- #2;
  \draw[line width=0.7pt,white] #1 -- #2;}

\providecommand{\scenesixdisc}[1]{%
  \tikz[baseline=(n.base)]{\node[circle,fill=s6mark,text=white,inner sep=0pt,
    minimum size=1.55ex,font=\sffamily\bfseries\tiny] (n) {#1};}}

\providecommand{\scenesixdiscsize}{1.5ex}
\tikzset{
  s6waypoint/.style={circle,fill=s6mark,draw=white,line width=0.5pt,text=white,
      inner sep=0pt,minimum size=\scenesixdiscsize,font=\sffamily\bfseries\tiny},
  s6badge/.style={circle,fill=s6mark,draw=white,line width=0.5pt,text=white,
      inner sep=0pt,minimum size=\scenesixdiscsize,font=\sffamily\bfseries\tiny},
  s6inset/.style={anchor=south west,inner sep=0pt,outer sep=0pt},
}

\providecommand{\scenesixlink}[3]{%
  \scenesixleader{#1}{#2}%
  \node[s6waypoint] at #1 {#3};%
  \node[s6badge]    at #2 {#3};}

\providecommand{\scenesixkey}{%
  {\sffamily\footnotesize\textcolor{black!65}{%
    \scenesixdisc{1}--\scenesixdisc{4} on the trajectory: where each on-board
    frame was taken}}}

\newcommand{\qualbluebincallouts}{%
  \node[s6inset] (T1) at (0.745,0.730)
    {\scenesixthumbmask{\scenesixthumbw}{rollout_onboard_snippets/blue_bin_obstacle_onboard_1.jpg}{\fill[s6maskA] (0.035,0.620) rectangle (0.180,0.700);}};
  \node[s6inset] (T2) at (0.515,0.730)
    {\scenesixthumb{\scenesixthumbw}{rollout_onboard_snippets/blue_bin_obstacle_onboard_2.jpg}};
  \node[s6inset] (T3) at (0.285,0.730)
    {\scenesixthumb{\scenesixthumbw}{rollout_onboard_snippets/blue_bin_obstacle_onboard_3.jpg}};
  \node[s6inset] (T4) at (0.055,0.730)
    {\scenesixthumb{\scenesixthumbw}{rollout_onboard_snippets/blue_bin_obstacle_onboard_4.jpg}};
  \scenesixlink{(0.950,0.508)}{(T1.south east)}{1}
  \scenesixlink{(0.789,0.451)}{(T2.south east)}{2}
  \scenesixlink{(0.640,0.372)}{(T3.south east)}{3}
  \scenesixlink{(0.480,0.392)}{(T4.south east)}{4}}

\newcommand{\qualyellowcallouts}{%
  \node[s6inset] (Y1) at (0.745,0.730)
    {\scenesixthumb{\scenesixthumbw}{rollout_onboard_snippets/yellow_signs_onboard_1.jpg}};
  \node[s6inset] (Y2) at (0.515,0.730)
    {\scenesixthumbmask{\scenesixthumbw}{rollout_onboard_snippets/yellow_signs_onboard_2.jpg}{\fill[s6maskB] (0.318,0.635) rectangle (0.430,0.690);}};
  \node[s6inset] (Y3) at (0.285,0.730)
    {\scenesixthumb{\scenesixthumbw}{rollout_onboard_snippets/yellow_signs_onboard_3.jpg}};
  \node[s6inset] (Y4) at (0.055,0.730)
    {\scenesixthumb{\scenesixthumbw}{rollout_onboard_snippets/yellow_signs_onboard_4.jpg}};
  \scenesixlink{(0.950,0.485)}{(Y1.south east)}{1}
  \scenesixlink{(0.781,0.486)}{(Y2.south east)}{2}
  \scenesixlink{(0.600,0.420)}{(Y3.south east)}{3}
  \scenesixlink{(0.400,0.398)}{(Y4.south east)}{4}}

\newcommand{\qualBlueBin}{%
  \scenesixannot[\scenesixcamw]{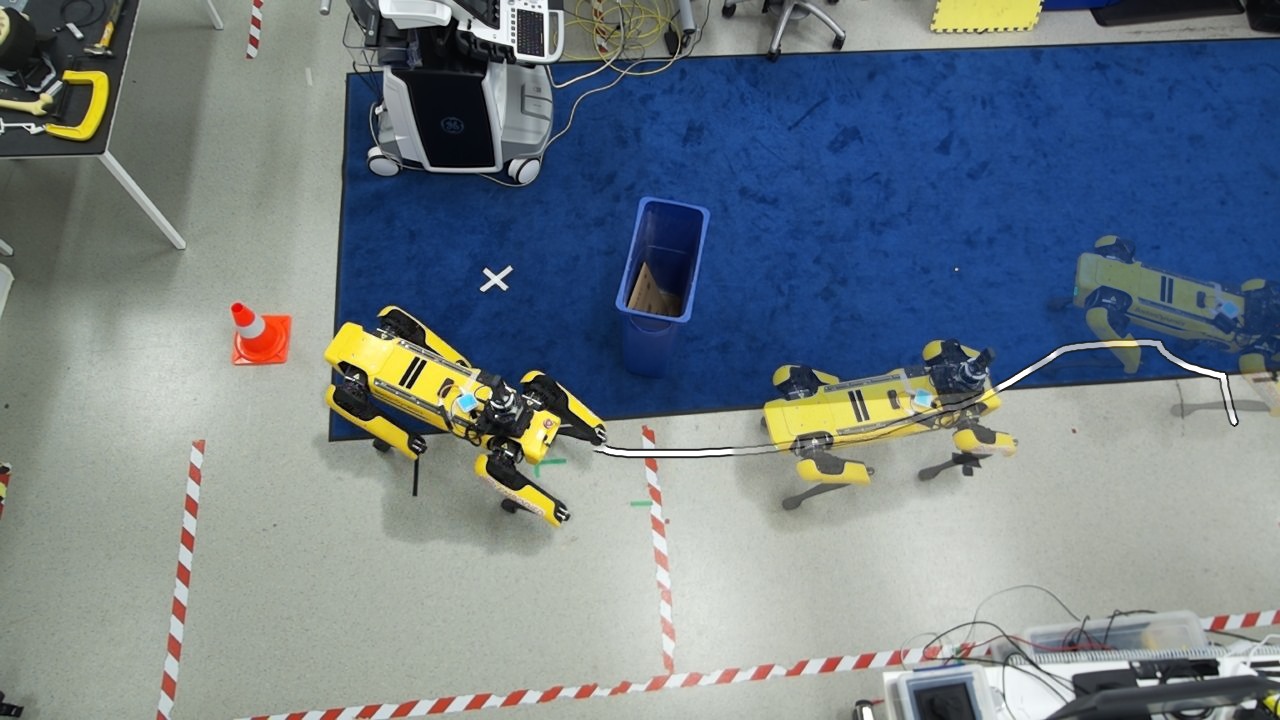}%
    {\qualbluebincallouts}}

\newcommand{\qualYellowSigns}{%
  \scenesixannot[\scenesixcamw]{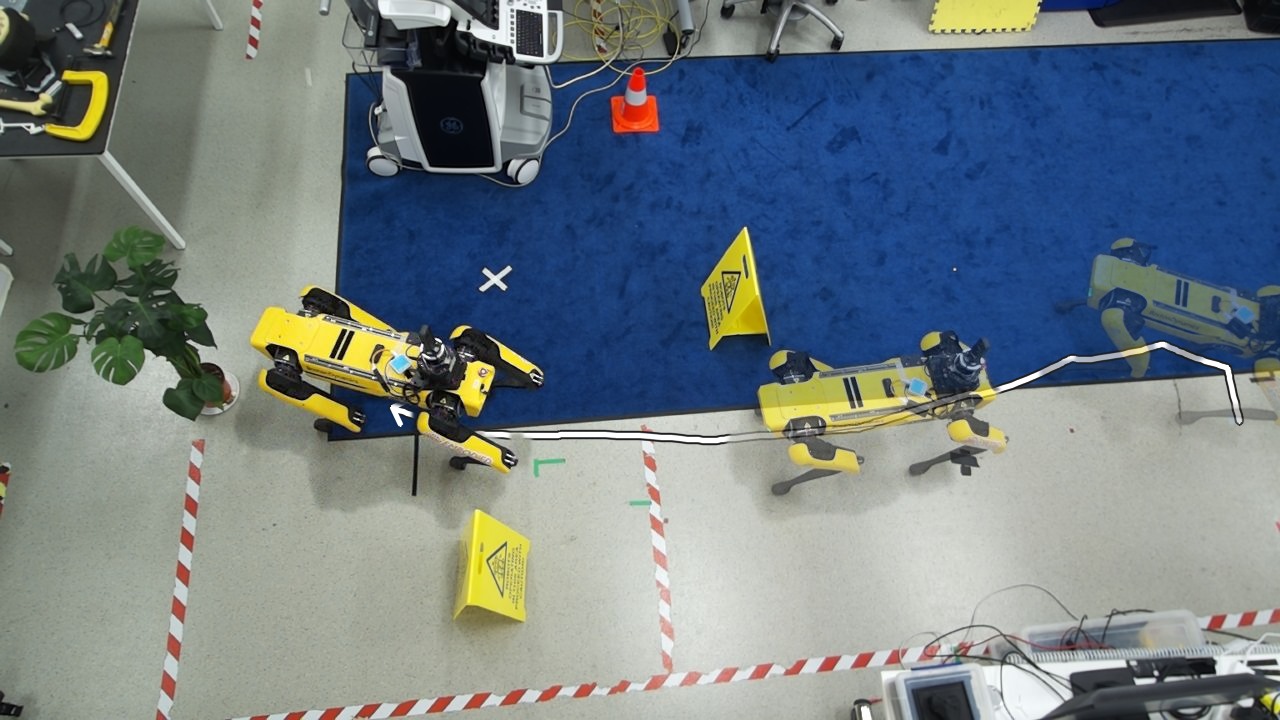}%
    {\qualyellowcallouts}}

\renewcommand{\scenesixcropbottom}{0.08132\scenesixcamw}
\setlength{\scenesixcamw}{\dimexpr0.82\columnwidth-0.8pt\relax}

\begin{figure}[t]
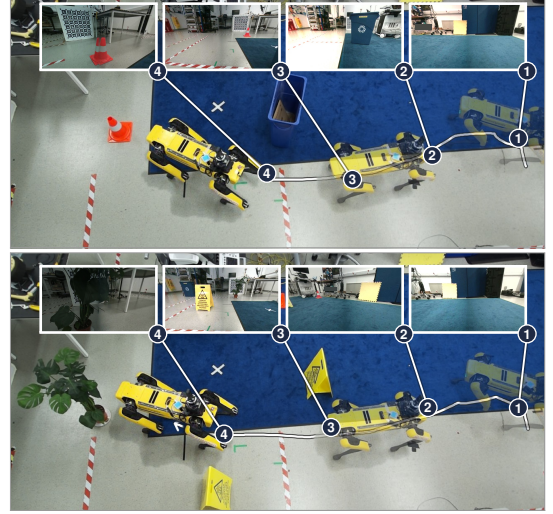

  \centering
  \qualBlueBin\\[1.5pt]
  \qualYellowSigns
  \caption{Obstacle avoidance preserved from pretraining, seen from \acs{cctv}~1. Top: instructed to reach the traffic cone, the policy walks around a blue bin. Bottom: instructed to reach the plant, it passes between two warning signs. The strip above each photograph holds four on-board frames in chronological order, each linked to the numbered disc marking roughly where it was taken; the disc positions are approximate. A logo in the background of two frames has been masked for anonymity.
  \vspace{-20pt}
  }
  \label{fig:qualitative}
\end{figure}

\textbf{Obstacle avoidance.} Two further rollouts test whether the policy keeps its pretrained obstacle avoidance (\cref{fig:qualitative}). We placed a bin between the robot and the cone and, in a second scene, two warning signs before the plant. Neither object appears in our demonstrations, which contain no obstacle avoidance and never place an obstacle in front of a target. In both rollouts, the policy turned toward the named object, walked around the obstacle, and reached it; with the bin, the cone stayed hidden even after the turn, a situation the training data does not cover either. These are two rollouts on one side of the area, so we report them as qualitative evidence, but they show what the combination makes possible: guided by the infrastructure cameras, the policy walks to an object it cannot see and, with what it learned in pretraining, around an obstacle it has never seen before.

\subsection{Ablations}
\label{sec:exp-ablations}
\begin{table}[t]
  \caption{Ablations of the architecture, on the semantic task with the stage-1 recipe, 50 rollouts each. First row: the stage-1 policy of \cref{tab:semantic}; every other row changes one decision of \cref{sec:method-architecture}. SR and clearance as in \cref{tab:semantic}.}
  \label{tab:ablation}
  \centering\footnotesize
  \begin{tabular}{@{}lcc@{}}
    \toprule
    Policy & SR (\%) $\uparrow$ & Clearance (m) $\uparrow$ \\
    \midrule
    \method, upsampling (stage 1) & \textbf{100.0} & 0.71 \\
    \midrule
    16 CCTV tokens per view    & 28.0 & \textbf{0.77} \\
    Shared CCTV encoder        & 96.0 & 0.61 \\
    CCTV tokens first          & 94.0 & 0.65 \\
    Frozen on-board encoder    & 90.0 & 0.59 \\
    Frozen on-board encoder and LLM & 26.0 & 0.61 \\
    \bottomrule
  \end{tabular}
  \vspace{-20pt}
\end{table}
All ablations use the semantic task of \cref{sec:exp-semantic} and are trained through stage 1 only, with the same recipe and data as our method (\cref{tab:ablation}). Pooling each view further drops performance to near chance. Neither freezing variant we tried helps: freezing the \ac{llm} along with the on-board encoder performs near chance, and freezing only the on-board encoder makes the robot drive into a shelf in five rollouts. A shared \ac{cctv} encoder or \ac{cctv} tokens inserted elsewhere also fail on a few rollouts, mostly to worse local navigation.

\section{Conclusion}
\label{sec:conclusion}
Many indoor environments already have cameras, but navigation policies do not use them. We presented \method, which extends a pretrained navigation \ac{vla} with static infrastructure views, together with a training recipe designed to leverage them. In two simulated warehouse tasks the policy used the cameras to identify objects named in the instruction and to reroute around blocked aisles, each time with information its own cameras could not provide. On a real robot, the same approach worked with less than 10 minutes of demonstrations and kept the obstacle avoidance of the pretrained model. These results show the benefit of augmenting a navigation \ac{vla} with sensing from the camera infrastructure that indoor environments already have.

\textbf{Limitations and future work.} Every result comes from a single training run, and several seeds would allow a statistical comparison. All experiments use one facility with at most four cameras, and the two language-instructed tasks start from a fixed pose, so larger buildings, more cameras, and arbitrary start poses remain untested. Inference time doubles to \qty{190}{\milli\second} with four added \ac{cctv} views and grows with their number, limiting scalability. Recovery data on the real robot remains untested, since mining from the policy's rollouts needs a larger area and more arrangements than ours.

\enlargethispage{4\baselineskip}
\section*{ACKNOWLEDGMENT}
Research supported by the NVIDIA Academic Grant Program using an NVIDIA RTX PRO 6000 Blackwell GPU and NVIDIA Isaac Sim, and by the EPSRC Programme Grant ``From Sensing to Collaboration'' (EP/V000748/1). Claude Code (Anthropic) assisted with the writing of this paper and the code.

\bibliographystyle{IEEEtran}
\bibliography{bstctl,references_compact}

\end{document}